\documentclass[journal]{IEEEtran}
\ifCLASSINFOpdf
\else
\fi

\usepackage{epsfig}
\usepackage{graphicx}
\usepackage{amsmath}
\usepackage{amssymb}
\usepackage{bm}
\usepackage{multirow}
\usepackage{booktabs}
\usepackage{enumerate}
\usepackage{hyperref}
\usepackage{changes}
\usepackage{float}
\hypersetup{colorlinks}  
\usepackage[section]{placeins}
\usepackage{caption}
\usepackage{subcaption}
\usepackage{soul, xcolor}
\usepackage{graphicx}

\makeatletter
\newif\if@restonecol
\makeatother

\usepackage[linesnumbered,ruled,vlined]{algorithm2e}
\usepackage{algpseudocode}
\usepackage{amsmath}

\begin{document}
%

\title{Automated Deepfake Detection}

\author{Ping Liu,
        Yuewei Lin,
        Yang He,
        Yunchao Wei,
        Liangli Zhen,
        Joey Tianyi Zhou,
        Rick Siow Mong Goh,
        Jingen Liu
\IEEEcompsocitemizethanks{\IEEEcompsocthanksitem P. Liu, Joey Zhou, L. Zhen, Rick Goh are with Institute of High Performance Computing, Agency for Science, Technology, and Research, Singapore. 
\IEEEcompsocthanksitem Y. He and Y. Wei are with Centre for Artificial Intelligence, University of Technology Sydney, Sydney, Australia. 
\IEEEcompsocthanksitem Y. Lin is with Brookhaven National Laboratory, Upton, NY, USA.
\IEEEcompsocthanksitem J. Liu is with JD AI Research, Mountain View, CA, USA.
\IEEEcompsocthanksitem  Joey Tianyi Zhou is the corresponding author.
}

\thanks{Manuscript received Jan 22, 2021.}}

\markboth{Journal of \LaTeX\ Class Files,~Vol.~14, No.~8, August~2015}%
{Shell \MakeLowercase{\textit{et al.}}: Bare Demo of IEEEtran.cls for Computer Society Journals}

\IEEEtitleabstractindextext{%
\begin{abstract}
In this paper, we propose to utilize Automated Machine Learning to adaptively search a neural architecture for deepfake detection. This is the first time to employ automated machine learning for deepfake detection. Based on our explored search space, our proposed method achieves competitive prediction accuracy compared to previous methods. To improve the generalizability of our method, especially when training data and testing data are manipulated by different methods, we propose a simple yet effective strategy in our network learning process: making it to estimate potential manipulation regions besides predicting the real/fake labels. Unlike previous works manually design neural networks, our method can relieve us from the high labor cost in network construction. More than that, compared to previous works, our method depends much less on prior knowledge, \textit{e.g.}, which manipulation method is utilized or where exactly the fake image is manipulated. Extensive experimental results on two benchmark datasets demonstrate the effectiveness of our proposed method for deepfake detection.
\end{abstract}

\begin{IEEEkeywords}
Deepfake Detection, Neural Architecture Search, Potential Manipulation Region Localization, Facial Activity Analysis.
\end{IEEEkeywords}

}

\maketitle

\IEEEdisplaynontitleabstractindextext

\IEEEpeerreviewmaketitle


\section{Introduction}\label{sec:introduction}

\IEEEPARstart{D}{e}ep fake detection aims to tell us whether the face in a given image is synthesized (fake) or not (real). With the development of {computer graphics and computer vision}~\cite{abdal2019image2stylegan,karras2020analyzing,wu2019relgan,DBLP:journals/corr/abs-2103-05193,9110728,9318504}, the generated faces become so visually realistic, resulting in the difficulty even for humans to differentiate. The unauthorized and malicious distribution of those fake images or videos brings serious concern among communities. To track the issue, an efficient and effective solution for deepfake detection becomes urgently required. In past three years, deep learning methods, \textit{e.g.}, convolutional neural networks (CNNs), have been utilized for general forgery image classification \cite{7086315_tifs2015,7154457_tifs2015,6987281_tifs2015} and deepfake detection \cite{9298826_tifs2021,9505637_tifs2021} due to their promising performance in various computer vision applications,~\textit{e.g.}, image classification~\cite{8253869, 9376703,9376704_heheefan_2021,8933048_tifs2020,9346018,luo2020every_pr2020}, image retrieval~\cite{Peng2021Joint, Miao_2019_ICCV,Zheng2021,8485427,9068282_tcsvt2021}, and semantic segmentation~\cite{9108530,luo2018macro, luo2019significance,luo2020ASM,pan2020adversarial}.

\begin{figure}[htbp]
  \centering
  \includegraphics[width=1.0\linewidth]{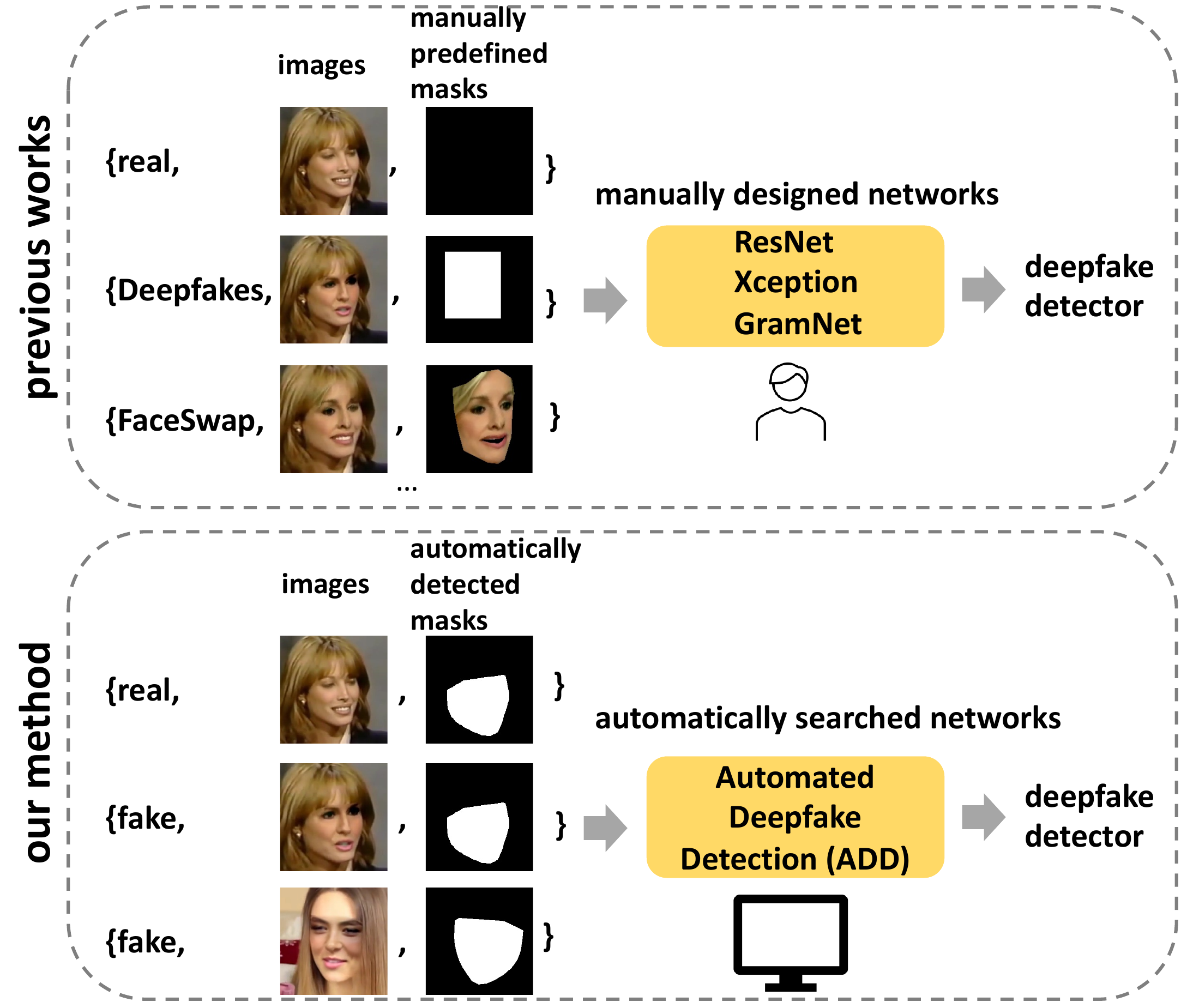}  
  \caption{ {The difference between previous deepfake detection methods and our automated deepfake detection (ADD).}}
  \label{fig:manual_auto}
\end{figure}

Most recent CNNs-based deepfake detection methods~\cite{he2016deep,chollet2017xception,Du2020} are built on backbones which are originally designed for other computer vision tasks, such as image classifications (ResNet~\cite{he2016deep}, XceptionNet~\cite{chollet2017xception}), or segmentation (U-Net~\cite{Du2020}). Directly borrowing the neural architectures from image classification tasks makes them fail to consider the specific characteristics of deepfake detection and might lead to inferior performance. For example, image classification tasks mainly focus on the shape, color, and semantic difference between different categories, while deepfake detection relies more on local texture discrepancies between different regions. 


Some works, such as~\cite{liu2020global,Zhou2021_cvpr2021}, manually designed neural architectures after analyzing the characteristics of deepfake detection tasks. Specifically, Liu \textit{et al.}~\cite{liu2020global} designed a new architecture named Gram-Net, making the network focus on texture discrepancies. This designed Gram-Net not only achieves promising performance in fake image detection but also facilitates further result interpretations. In \cite{Zhou2021_cvpr2021}, Zhou \textit{et al.} manually designed a discriminative attention model for detecting forgery faces in a multi-person scenario. However, as pointed out in previous works~\cite{Dong_2019_CVPR}, manually design neural architectures for specific tasks is a trial-and-error process, which is intractable  and time-consuming.


To handle aforementioned limitations in previous deepfake detection works such as \cite{ liu2020global}, we propose a method, which is called Automated Deepfake Detection (ADD), to construct a neural architecture for deepfake detection in an adaptive and automatic manner. As shown in Fig.\ref{fig:manual_auto}, benefiting from automated machine learning (AutoML)~\cite{liu2018darts, Dong_2019_CVPR}, ADD is able to relieve us from the heavy labor cost during manual network design processes, while still assists us in achieving a good balance between prediction accuracy and model sizes. We use our method to search neural cells, stack them hierarchically to build a deep neural network for deepfake detection. To the best of our knowledge, it is the first time to propose an AutoML based framework for deepfake detection. 

Moreover, to further improve the generalizability of our constructed neural network for deepfake detection, we additionally introduce a simple yet effective strategy to our ADD learning process: localizing the \textit{potential} manipulation region. Here, being ``potential" means our method does not require prior knowledge such as what manipulation method is applied \cite{Du2020,Mazaheri2021}. By utilizing this strategy, our ADD learns two tasks simultaneously,~\textit{i.e.}, differentiating fake samples from real ones as well as locating \textit{potential} manipulated regions (even no manipulation occurs). Explicitly locating potential manipulation regions has two advantages. On the one hand, it makes the network focus on features extracted in those potentially manipulated regions and depresses the irrelevant information, relieving the networks from disruptions; on the other hand, explicitly localizing the manipulated regions benefits the downstream interpretation. To segment \textit{manipulated} regions for a similar purpose, previous works such as~\cite{Du2020,Mazaheri2021} depend heavily on various prior knowledge, such as which manipulation method is applied, which part is manipulated in each fake sample, and etc. In contrast, our method can work with few dependence on prior knowledge. Our experimental results empirically prove that our ADD equipped with the proposed potential manipulation region localization strategy can bring significant performance improvement in cross-dataset evaluations (Table \ref{tab:cross-dataset_face2face_faceswap} and \ref{tab:cross-dataset_ffpp_Celeb-DF}).

In summary, our contributions are listed as follows:

(1) This is the first time to apply automated machine learning to search neural architectures for deepfake detection. Compared to previous works manually design deep networks or borrow architectures designed for other tasks, our method can achieve high prediction accuracy based on automatically searched architectures. Besides that, we propose a potential manipulation region detection strategy that is simple yet effective to boost the generalization ability of our method. Specifically, our method not only predicts the real or fake label for each given sample but also locates the potential manipulation region with few dependence on prior knowledge. 



(2) We conduct extensive experiments on two challenging datasets about deepfake detection,~\textit{i.e.}, FaceForensics++ (FF++)~\cite{Rossler2019} and Celeb-DF~\cite{Li2020_cvpr2020}. On both inner-dataset and cross-dataset settings, our method achieves a competitive performance comparing to previous works.



\section{Related Work}
In this section, we briefly review previous works related to our work, including deepfake detection and neural architecture search.

\subsection{Deepfake Detection}
Deepfake is a kind of synthetic media, where the human face in a source image is replaced with a different face provided by a target image, or the face attribute in a source image is replaced with that in a target image. By leveraging the latest computer graphics technologies and generative machine learning methods~\cite{goodfellow2014generative_nips,yeh2016semantic,ding2018exprgan_aaai,park2020swapping_nips,hu2020unsupervised,shen2020interpreting_cvpr}, the manipulated images have nearly no visual clues, making it hard to distinguish them from real images. In past years, a variety of works for deepfake detection have been proposed, most of which treat deepfake detection as merely a binary image classification problem (real or fake). For instance, as an initial attempt,~\cite{Cozzolino2017,fridrich2012rich} utilize handcrafted features and steganalysis to construct a binary classifier for deepfake detection. With the advances of CNNs, researchers leverage manually designed neural architectures or neural cells to build networks for deepfake detection~\cite{Rossler2019,Afchar2018,Zhou2017,Nguyen2019}. Some of previous deep learning based works, such as~\cite{Rossler2019,Afchar2018,Zhou2017,Nguyen2019}, directly apply neural architectures manually designed for natural image classification.~\cite{Afchar2018} builds a CNN based on inception modules \cite{szegedy2015going_cvpr} and trains the constructed network under a supervision of a mean squared error loss.~\cite{Rossler2019} transfers XceptionNet~\cite{chollet2017xception} from natural image classification to deepfake detection by modifying the output layer.~\cite{Zhou2017} constructs a two-branch structure for deepfake detection, one of which is a GoogLeNet~\cite{szegedy2016rethinking} pretrained on ImageNet.~\cite{Nguyen2019} presents a method based on Capsule network~\cite{Sabour2017}. ~\cite{Rossler2019,Afchar2018,Zhou2017,Nguyen2019} treat deepfake detection as a vanilla binary classification problem, with few considerations about the differences between natural image classification and deepfake detection. For example, comparing to natural image classifications, deepfake detection focuses on human faces, which are with strict topological structures.

From a different perspective,~\cite{Du2020,Mazaheri2021,Wang2020,Li2020,Nirkin2021} analyze the characteristics of deepfake images and introduce specific prior knowledge into their solutions. In~\cite{Du2020,Mazaheri2021,Wang2020}, besides predicting real or fake for given samples, the position of the manipulated region in each fake sample is also estimated simultaneously. Treating the manipulated region localization as a segmentation task,~\cite{Du2020,Mazaheri2021,Wang2020} build their network on the architectures originally designed for segmentation, such as U-Net~\cite{Du2020}, Encoder-Decoder structure~\cite{Mazaheri2021,Nirkin2021}.~\cite{Li2020_cvpr2020} utilizes XceptionNet equipped with an additional segmentation branch. Li \textit{et al.} \cite{Li2020_cvpr2020} argue that there is a blending boundary existing in fake samples if a Poisson blending is utilized to post-process the manipulated face images. They name the blending boundary as face X-ray,  whose existence informs us the given sample is manipulated and whose position indicates where it is manipulated. Comparing to prior works~\cite{Du2020,Mazaheri2021,Wang2020,Li2020}, Li \textit{et al.}~\cite{Li2020_cvpr2020} require external data collection and annotation to train their model, since there are no such annotations for detecting face X-ray previously. Most of the previous works on deepfake detection only utilize spatial domain information for training their network. Recently, works such as \cite{Qian2020_eccv2020, Shen2021_aaai2021,Li2021_cvpr2021} manually design networks to mine forgery patterns from a frequency domain. They believe that the frequency domain can provide complementary knowledge to detect forgery patterns in manipulated face images, which has been experimentally proved in \cite{Qian2020_eccv2020, Shen2021_aaai2021,Li2021_cvpr2021,Frank2020_icml2020, Liu2021_cvpr2021}. For a systematic review for deepfake generation and detection, please refer to~\cite{Mirsky2020}.

Although some similarities might exist between~\cite{Du2020,Mazaheri2021} and our work, there are significant differences between their proposed methods and ours: (1)~\cite{Du2020,Mazaheri2021} directly borrow networks manually designed for image classification or segmentation, while our method utilizes a gradient update method to automatically search an architecture for deepfake detection. Based on our experimental observations (Fig. \ref{fig:searched_cell}), the searched architecture is significantly different from that in previous works and is more adaptive to achieve competitive performance; {(2)}~\cite{Du2020,Mazaheri2021} require strong prior knowledge for detecting the manipulation regions. Concretely, it is necessary to know which manipulation method is utilized to generate fake images in \cite{Du2020,Mazaheri2021}. Introducing incorrect prior knowledge might deteriorate the prediction accuracy since different manipulation methods correspond to different manipulated regions. For example, the manipulated region is around the whole face when conducting FaceSwap \cite{FaceSwap}; when NeuralTexture \cite{Thies2019} is utilized to manipulate face images, the manipulated regions are around the mouth only. The heavy dependence on prior knowledge limits practical applications of~\cite{Du2020,Mazaheri2021}, since in real scenarios it is hard to know which manipulation method is utilized. In our experiment, it is proved that the incorrect localization about manipulation regions has a negative impact, especially when testing samples and training samples are manipulated on different regions (Table \ref{tab:cross-dataset_face2face_faceswap}). In contrast, our method does not require that prior knowledge as~\cite {Du2020,Mazaheri2021} do since we focus on detecting \textit{potential} manipulation regions.  

\subsection{Automated Machine Learning}
Automated Machine Learning (AutoML) focuses on automating various aspects in machine learning tasks, such as neural architecture construction \cite{Dong_2019_CVPR}, hyperparameter setting \cite{he2018amc_eccv}, data augmentation policies \cite{cubuk2019autoaugment_cvpr}, and etc. Neural Architecture Search (NAS), as one sub-research direction of AutoML, aims to automatically search a neural architecture based on given tasks and data. NAS is expected to decrease the dependence on human expertise during the whole network design process~\cite{Li2020, Du2020,Wang2020,Mazaheri2021}. In past years, the efficacy of NAS has been proved in various tasks,~\textit{e.g.}, image classification~\cite{real2019regularized}, semantic segmentation~\cite{liu2019auto}, object detection~\cite{ghiasi2019fpn}, pose estimation~\cite{zhang2020efficientpose}, image super-resolution~\cite{song2020efficient}, model compression \cite{he2018amc_eccv,dong2019one_iccv2019,dong2019network_neurips2019,He_2020_CVPR}, and etc. Generally, NAS conducts the architecture search in an iterative and alternative manner. Before starting the search, a search space consisting of pre-defined candidates is provided. In each search iteration, promising candidates are selected and updated for the following iterations, while poor candidates are dropped without further consideration. Fixing the architecture constructed by those selected candidates, the parameters of the searched candidates are learning in an alternative manner. The search and update operation continues iteratively until a satisfactory performance is achieved. Previous NAS methods can be categorized into two major groups: reinforcement learning-based methods such as~\cite{zoph2018learning}, and gradient-based methods such as~\cite{Dong_2019_CVPR,liu2018darts}. In reinforcement learning-based works like~\cite{zoph2018learning}, reinforcement learning algorithms are leveraged to decide whether the searched candidates are on the correct track~\cite{zoph2018learning}.  Although reinforcement learning-based methods have been proved effective in a variety of works~\cite{zoph2018learning}, the training difficulty and huge computation cost limit their further applications. To solve those limitations, ~\cite{Dong_2019_CVPR,liu2018darts} propose gradient update methods to speed up the neural architecture search process. The basic idea of gradient-based NAS works~\cite{Dong_2019_CVPR,liu2018darts} is to utilize a continuous relaxation for the architecture representation, making the learning process can be updated using gradient descent methods. Compared to the reinforcement learning-based methods~\cite{zoph2018learning}, gradient update based methods~\cite{Dong_2019_CVPR,liu2018darts} can conduct the whole search process in a more efficient and stable manner. 

\section{Methodology}
\label{sec:method}
In this section, we describe the technical details of our automated deepfake detection method.  We first introduce the preliminary background of NAS in Sec.~\ref{subsec:preli}; at second, in Sec.~\ref{subsec:sdd}, we illustrate the search space utilized in our automated deepfake detection method; at third, we design our learning strategy, \textit{i.e.}, differentiating real samples from fake samples and locating {potential} manipulation regions simultaneously.

\subsection{Preliminaries}
\label{subsec:preli}
In this work, we follow previous NAS works \cite{Dong_2019_CVPR,liu2018darts} and conduct our neural architecture search in a micro-search manner. The target of micro-search works \cite{Dong_2019_CVPR,liu2018darts}, is to automatically discover a neural cell and hierarchically stack this searched neural cell based on a pre-defined structure. Compared to its counterpart, \textit{i.e.}, the macro-search methods which directly discover the entire network architecture, the micro-search methods can reduce the search cost significantly. 

In micro-search methods \cite{Dong_2019_CVPR,liu2018darts}, the searched cell is viewed as a directed acyclic graph (DAG), in which each node is a block of operations. Each block receives two feature tensors as its input, processes them by two selected operations respectively, and then sums these two processed tensors. The operations to process the feature tensors are selected from a pre-defined operation set, which is known as \textit{search space}. The search space includes a variety of pre-specified operations, such as pooling (average/max) operation, convolution with different kernel sizes (\textit{e.g.}, $3 \times 3$, $5 \times 5$), identity mapping, and etc. Concretely, we denote $I_{i,1}^{c}$ and $I_{i,2}^{c}$ as the two input tensors, and $\mathcal{OP}_{i,1}^{c}(\cdot)$ and $\mathcal{OP}_{i,2}^{c}(\cdot)$ as the two corresponding selected operations in the $i_{th}$ block of the $c_{th}$ cell, then the output tensor $O_{i}^{c}$ is defined as:
\begin{equation}
    O_{i}^{c} = \mathcal{OP}_{i,1}^{c}(I_{i,1}^{c})+\mathcal{OP}_{i,2}^{c}(I_{i,2}^{c}).
\end{equation}
It is noted that $I_{i,1}^{c}$ and $I_{i,2}^{c}$ are output tensors from the adjacent stacked neural cells, \textit{i.e.}, {$O_{*}^{c-1}$ and $O_{*}^{c-2}$.}


\begin{figure*}[htbp]
  \centering
  \includegraphics[width=1\linewidth]{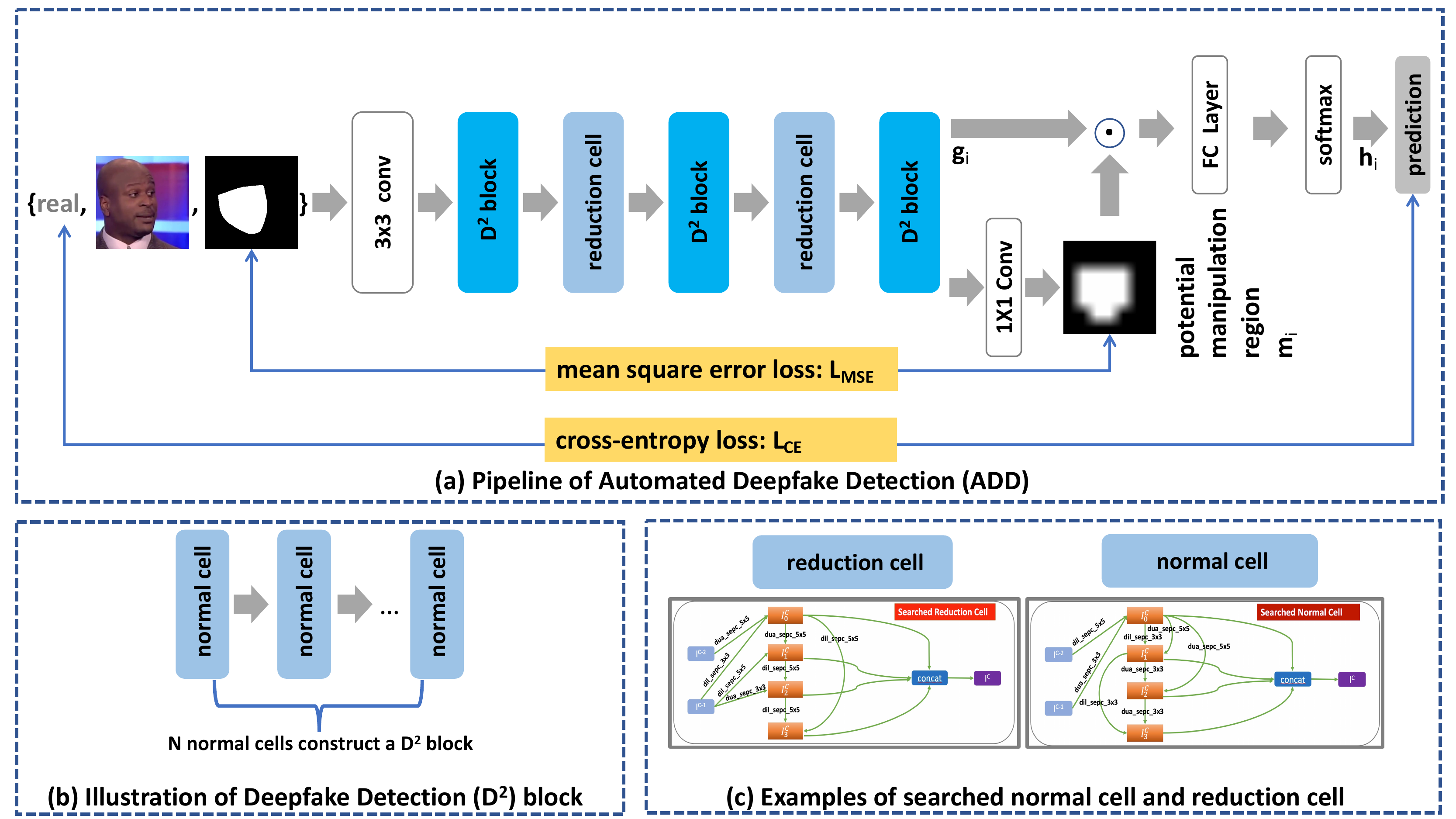}  
  \caption{ {Overview of our proposed automated deepfake detection method. (a) The pipeline of automated deepfake detection network. The network consists of deepfake detector ($D^2$) blocks and reduction cells. (b) Illustration of deepfake detector ($D^2$) blocks, each of which consists of N normal cells. (c). Examples of normal cells and reduction cells searched by our method.}}
  \label{fig:pipeline_add}
\end{figure*}

The key of neural architecture search is how to define a feasible search space based on target tasks, and how to efficiently select an optimal set of operations,~\textit{i.e.}, $\mathcal{OP}_{i,1}^{c}(\cdot)$ and  $\mathcal{OP}_{i,2}^{c}(\cdot)$ from the search space. The selected operations are leveraged to process two input feature tensors for generating an output feature tensor. The generated output feature tensor is utilized as an input tensor for the following block. As the operation selection is a categorical process in principle, previous works such as \cite{zoph2018learning}, utilize reinforcement learning to conduct the selection. To decrease the computational cost in reinforcement learning methods, we choose a more efficient strategy by following \cite{Dong_2019_CVPR,liu2018darts}: relaxing the categorical choice for a particular operation as a softmax process over all possible operations, which is defined as:
\begin{equation}
     \mathcal{OP}_{i,1}^{c}(I_{i,1}^{c})=\sum\limits_{H \in I_{i}^{c}} \sum\limits_{op \in \mathcal{OP}}\frac{\exp(\alpha_{op}^{H,i})}{\sum_{op' \in \mathcal{OP}}\exp(\alpha_{op'}^{H,i})} op(H)
\end{equation}
where  architecture parameter $\alpha=\{\alpha_{op}^{H,i}\}$ represents the topology structure for the searched cell. The cell topology structure indicates which operations are selected in the corresponding block. For all selected operations, their parameters are denoted as $w$. Therefore, in the neural architecture searching process, there are two groups of parameters that need to be learned: architecture parameter $\alpha$ and operation parameter $w$. The learning process is supervised by a loss function, which is to minimize the discrepancy between the ground truth and prediction:
\begin{eqnarray}
 \mathcal{L}(y;{\hat{y}})=\mathcal{L}(y; F({x}, \alpha, w)) 
     \label{eq:training_alpha_w}
\end{eqnarray}
in which $\mathcal{L}$ denotes a discrepancy measure function (\textit{e.g.}, cross-entropy), $y$ denotes ground truth labels,  $\hat{y}$ denotes predictions, {$x$} denotes data samples,   $\alpha$ denotes the architecture parameters, $w$ denotes the associated weights corresponding to $\alpha$, $F$ denotes the network used to estimate $\hat{y}$ based on {$x$}, $\alpha$, and $w$.

By the relaxation strategy proposed in \cite{Dong_2019_CVPR,liu2018darts}, the architecture parameter $\alpha$ and operation parameter $w$ can be learned in an alternative manner. Specifically, the architecture parameter $\alpha$ is trained on a validation set while the operation parameter $w$ is trained on a training set. During the training, the two sets of parameters can be updated by gradient descent, which runs in a high efficiency. After training, the candidate in $I_{i,*}^{c}$ with the maximum value, \textit{i.e.}, $\max \frac{\exp{\alpha_{op}^{H,i}}}{\sum \exp(\alpha_{op'}^{H,i})}$, is selected as $I _{i1}^{c}$, and the operation with the maximum weight is selected as $\mathcal{OP}_{i1}^{c}(\cdot)$. The whole training process is shown in Algorithm.\ref{alg:search_optimization}.

\vspace{-3mm}
\begin{algorithm}[!t]
  \caption{Searching Algorithm}
  \label{alg:search_optimization}
  \KwIn{a training set $D_{train}$, a validation set $D_{val}$,\\ randomly initialized $\alpha$, and random initialized $w$}

  \KwOut{optimized parameter $\alpha^{*}$ and $w^{*}$}
  
  \While {not converged  }
  
      \quad sample a batch of data from the training set, denoted as $d_{train,b}$, where $b$ is the batch index; \\
      \quad calculate loss $\mathcal{L}_{T}$ based on Eq. \ref{eq:training_alpha_w} for $d_{train,b}$; \\
      \quad update operation parameter $w$ by gradient descent: $w=w-\bigtriangledown_{w}\mathcal{L}_{T}$;\\
      \quad sample a batch of data from the validation set, denoted as $d_{val,b}$, where $b$ is the batch index; \\
        \quad calculate loss $\mathcal{L}_{V}$ based on Eq. \ref{eq:training_alpha_w} for $d_{val,b}$; \\
      \quad update architecture parameter $\alpha$ by gradient descent: $\alpha = \alpha - \bigtriangledown_{\alpha}\mathcal{L}_{V}$
\end{algorithm}
\vspace{-3mm}

\subsection{Search for Deepfake Detection}
\label{subsec:sdd}

\noindent
\textbf{Search Cells From Search Space} As a vital part in NAS, a search space consists of all possible operation candidates to be found. In previous works such as \cite{Dong_2019_CVPR,zoph2018learning}, they search neural architectures for natural image classification based on a search space containing different kinds of convolution layers, pooling layers, identity mapping, and etc. In this work, we conduct an exploration to define a search space for the deepfake detection problem.

Considering the similarity and difference between deepfake detection and other tasks, \textit{e.g.}, natural image classification, semantic segmentation, we design a deepfake search space $\mathcal{OP}_{df}$. At first, since deepfake detection is a classification problem, we introduce differnt kinds of vanilla convolution operations and pooling operations into $\mathcal{OP}_{df}$, since they have been proved effective in manually designed neural networks for deepfake detection \cite{afchar2018mesonet}; at second, since networks constructed by separable convolution layers \cite{Qian2020,Rossler2019} achieved promising results in deepfake detection, we also bring kinds of depth-wise separable convolution in $\mathcal{OP}_{df}$; at third, in deepfake detection, since the feature discrepancy between different local regions provides key clues for detecting forgery patterns, it is important to utilize certain operations for achieving a large receptive field. For this purpose, we introduce kinds of dilated convolution operations and large kernel convolution operations ($7 \times 7$) into our search space. In total, the search space $\mathcal{OP}_{df}$ for deepfake detection includes:
\begin{itemize}
  \item zero operation; 
  \item identity mapping; 
  \item $3\times 3$ average pooling;
  \item $3\times 3$ max pooling;
  \item $3\times 3$ convolution;
  \item $5\times 5$ convolution;  
  \item $7\times 7$ convolution;   
  \item $3\times 3$ separable convolution;   
  \item $5\times 5$ separable convolution; 
  \item $3\times 3$ dilated separable convolution;   
  \item $5\times 5$ dilated separable convolution. 
\end{itemize}

\textbf{Construct Network from Searched Cells} Two kinds of cells are searched based on our defined search space $\mathcal{OP}_{df}$: one is a normal cell, and the other one is a reduction cell. Each cell consists of operations selected from $\mathcal{OP}_{df}$. In the normal cell, the stride is set as $1$; in the reduction cell, the stride is set as $2$. After discovering one normal cell and one reduction cell, we stack $N$ normal cells as an operation block named Deepfake Detection ($D^2$) block . The reduction cell is placed between two $D^2$ blocks. 



\textbf{Learning by the Constructed Network} For sample $x_i$, we feed it to the constructed neural network and denote the extracted feature from the highest layer as $\textbf{f}_i \in \mathbb{R}^{c \times w' \times h'}$, where $c$ denotes the feature channel number, $w'$ and $h'$ denote the spatial size of the feature. The extracted feature $\textbf{f}_{i}$ is feed to a $1 \times 1$ convolution layer followed by two full connection layers (FC) to generate logits, which is denoted as ${g}_i \in \mathbb{R}^2$.

Given the calculated logits $g_{i}$, our constructed neural network is trained under a cross-entropy (CE) loss defined as:
\begin{eqnarray}
    \mathcal{L}_{CE} &=& \frac{1}{N}\sum_{i}cross\_entropy(y_i; FC(\textbf{f}_i) ) \nonumber \\
    &=&\frac{1}{N}\sum_{i}cross\_entropy(y_i; F(x_i, \alpha, w))
    \label{eq:lce}
\end{eqnarray}
where $N$ is the sample number, $y_i$ is the ground truth label for sample ${x}_i$, $F$ denotes the neural network constructed by our searched cells defined by parameters of  $\alpha$ and $w$.



\subsection{Potential Manipulation Region Localization}
In deepfake detection, the manipulated regions are located on faces, implying that the key information relevant to deepfake detection should be located on face regions rather than the background. More than that, previous face analysis works such as \cite{Liu2014, Zhang2020_ijcai} point out that not all facial regions make the contribution to the target task. As a matter of fact, some facial regions might make more contributions than other regions for the target task. Therefore, utilize features extracted from all image regions without different focus might be sub-optimal.

To solve this problem, we propose to make our searched neural architecture focus on the important regions for deepfake detection during the learning process. Specifically, for a sample $x_i$, we feed its extracted feature $\textbf{f}_{i}$ to another $1 \times 1$ convolution layer for generating a one channel feature map. This one channel feature map is feed to a bi-linear up-sampling layer for generating a mask $m_i \in \mathbb{R}^{1 \times w \times h}$. The spatial size of the mask $m_{i}$ is the same as that of the input ${x}_{i}$. Since we want our network to focus on the potential manipulation regions in each image, specifically, the face regions, we make the predicted $m_{i}$ have high activation values inside face regions and lower values outside non-face regions. We propose a simple and effective method to generate the ground truth $M_i$ for potential manipulation regions. Concretely, we use a light-weighted face detector to detect the face in the sample ${x}_{i}$, and detect facial keypoints on the face. The detected keypoints are utilized to generate a convex hull.  The ground truth $M_{i}$ is generated by setting all the pixels inside the convex hull as one and all the pixels outside the convex hull as zero. For making our neural network localize the clues related to forgery patterns, we train our model by additionally minimizing the discrepancy between $m_i$ and $M_i$ with a Mean Square Error (MSE) loss:
\begin{equation}
    \mathcal{L}_{MSE} = \frac{1}{N}\sum_{i}L_{2}(M_i; m_i)
    \label{eq:lmse_mask}
\end{equation}
where $m_i$ denotes the predicted potential manipulation region for the sample $x_i$, $M_i$ is the ground truth, and $N$ is the sample number.




In our method, the proposed potential manipulation region detection strategy is significantly different from the previous works like \cite{Mazaheri2021,li2020face_cvpr2020}, which explicitly detects the exact manipulated regions based on prior knowledge, such as which manipulation method is applied. Specifically, the meaning of the predicted potential manipulation region $m_i$ in our work is different from that of the estimated manipulated region, denoted as $s_{i}$, in previous works like \cite{Mazaheri2021}. In \cite{Mazaheri2021}, the value of each pixel in $s_{i}$ is between $0$ and $1$, indicating \textit{whether} or not a manipulation exists in the sample. In other words, the values of the mask $s_{i}$ indicate whether manipulation is applied to the corresponding sample. Other than that, the spatial range of $s_{i}$ varies according to the applied manipulation method. For example, in Deepfakes~\cite{Deepfakes} and FaceSwap~\cite{FaceSwap},  where the whole face in the source image is manipulated, $s_{i}$ covers the whole face region, while in NeuralTexture~\cite{Thies2019} where expressions are edited, $s_{i}$ only occupies the mouth region. Providing incorrect $s_{i}$ might cause the network to focus on wrong face regions to make an incorrect prediction, which happens if testing forgery face images are manipulated by a different method from the training fake images. In contrast, our method detects the potential manipulation regions, which does not need to know whether a manipulation method has been applied already or which manipulation method is applied. Few dependence on prior knowledge make our method with better generalizability, which is experimentally proved (Table \ref{tab:cross-dataset_face2face_faceswap} and \ref{tab:cross-dataset_ffpp_Celeb-DF}).


Correspondingly, with the predicted $m_{i}$, the loss term in Eq. \ref{eq:lce} is changed to:
\begin{eqnarray}
    \mathcal{L}_{CE'}=\frac{1}{N}\sum_{i}cross\_ entropy( y_i; FC(\textbf{f}_i\odot m_i)) 
    \label{eq:lce_mask}
\end{eqnarray}
where $\odot$ indicates an element-wise multiplication, $\textbf{f}_i$ denotes the extracted feature for sample $x_i$, $m_i$ denotes the estimated potential manipulation region in sample $x_i$, $y_i$ is the ground truth label, and $N$ is the sample number.

\subsection{Objective Function and Overall Algorithm}
The overall loss function to learn our automated deepfake detector is defined as follows:
\begin{eqnarray}
    \mathcal{L}_{overall}=\mathcal{L}_{CE'} + \alpha * \mathcal{L}_{MSE}
    \label{eq:l_overall}
\end{eqnarray}
where $L_{CE'}$ is defined in Eq. \ref{eq:lce_mask}, $L_{MSE}$ is defined in Eq. \ref{eq:lmse_mask}, $\alpha$ is a parameter to balance contributions of the two loss terms. The framework is trained in an end-to-end manner based on gradient descent strategy. The implementation details can be found in Section. \ref{subsec:imple}. The overall pipeline is shown in Fig. \ref{fig:pipeline_add}.

\section{Experiments}
\label{section:experiments_results}
In this section, we evaluate our proposed method on two deepfake detection datasets. We first illustrate the dataset details in Sec.~\ref{subsec:datasets}, experimental settings in Sec. \ref{subsec:imple}.  In Sec.~\ref{subsec:comparison}, we evaluate our method against previous works on two benchmark datasets. The ablation study is conducted in Sec.~\ref{subsec:ablation}. In Sec. \ref{subsec:visualization}, the interpretability of our method via visualization analysis is provided. In Sec. \ref{subsec:discussions} and Sec. \ref{subsec:limitations} we provide a discussion and the limitations of our method.

\subsection{Datasets}
\label{subsec:datasets}
Following previous works, We test our method on two recently released benchmark datasets: FaceForensics++~\cite{Rossler2019} (FF++), Celeb-DF~\cite{Li2020_cvpr2020}.

\textbf{FaceForensics++}~\cite{Rossler2019} includes four different types of face manipulation methods: Face2Face~\cite{MatthiasNiessner2016}, NeuralTextures~\cite{Thies2019}, FaceSwap~\cite{FaceSwap}, and Deepfakes~\cite{Deepfakes}. For each manipulation method, there are $1,000$ real videos and $1,000$ fake videos, which are split into $720$ videos for training, $140$ videos for validation, and $140$ videos for testing. In the four manipulation methods, Face2Face~\cite{MatthiasNiessner2016} and FaceSwap~\cite{FaceSwap} are computer graphics-based methods, NeuralTextures~\cite{Thies2019} and Deepfakes~\cite{Deepfakes} are learning-based approaches.

\begin{itemize}
  \item Face2Face~\cite{MatthiasNiessner2016} proposes a facial reenactment method to transfer expressions from a source video to a target video without changing the identity of the person in target videos. 
  \item NeuralTextures~\cite{Thies2019} utilizes a rendering approach to conduct facial reenactments. It uses a photometric reconstruction loss as well as an adversarial loss to train a rendering network. The trained network modifies the facial expression of the person in a target video. It should be noted that in NeuralTextures only the mouth region is modified, while other regions remain unchanged.
  \item FaceSwap~\cite{FaceSwap} transfers the face region from a source video to the corresponding region in a target video. At first, FaceSwap fits a 3D template model based on extracted facial landmarks in a source video; then, it back-projects the model to the target image by minimizing the shape difference. To make the result realistic, image blending and color correction are utilized.
  \item Deepfakes~\cite{Deepfakes} conducts face replacement based on deep learning, in which an encoder-decoder structure is utilized. An encoder and a decoder, which are trained to reconstruct training images of the source videos, are applied to the target faces. A Poisson blending process is utilized to post-process the output. 
 
\end{itemize}
There are three different video quality levels in FF++ dataset, which include: Low Quality (LQ), High Quality (HQ), and RAW, respectively. High quality means the data is generated by a constant rate quantization parameter of $23$, low quality means the data is generated by a quantization level of $40$. Unless stated, the data utilized in our experiment is in HQ format. In ablation studies, we also conduct an experiment on Deepfakes~\cite{Deepfakes} in LQ format to investigate the robustness of our method.

\textbf{Celeb-DF}~\cite{Li2020_cvpr2020} is a latest released dataset with improved visual qualities. The Celeb-DF dataset has $590$ real videos and $5,639$ fake videos in total. The real videos are collected from YouTube, which has $59$ different celebrities. It is generated using advanced deepfake synthesis techniques. The generated results are post-processed to correct color mismatch, temporal flickering, making the generated results with better visual qualities.

\begin{figure*}[htbp]
\begin{center}
\includegraphics[width=1.0\linewidth]{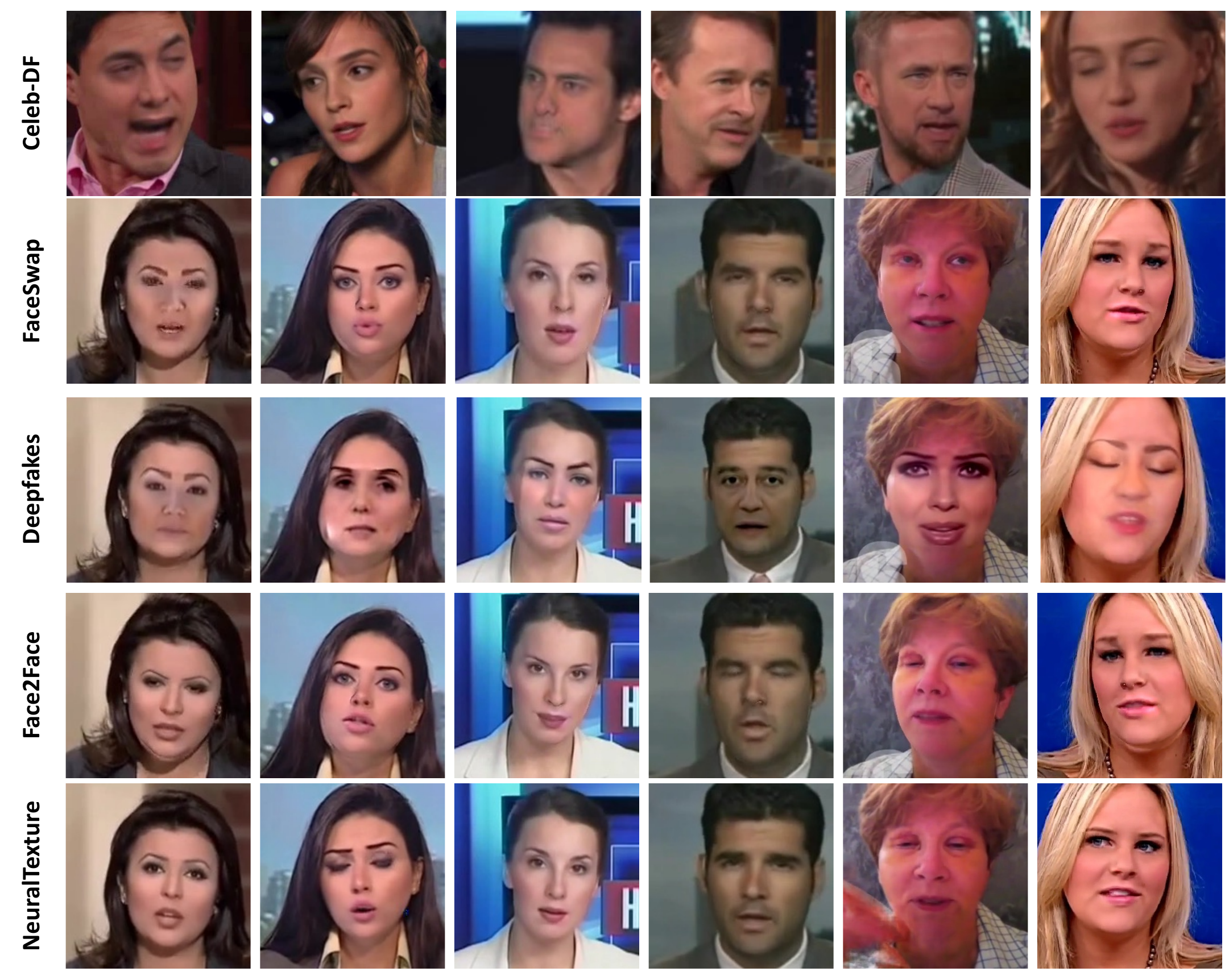}
\end{center}
\caption{Examples of Celeb-DF~\cite{Li2020_cvpr2020}, FaceSwap~\cite{FaceSwap}, Deepfakes~\cite{Deepfakes}, Face2Face~\cite{MatthiasNiessner2016}, and NeuralTextures~\cite{Thies2019} . }
\label{fig:examples_ff_Celeb-DF}
\end{figure*}

Some examples from Celeb-DF~\cite{Li2020_cvpr2020}, Face2Face~\cite{MatthiasNiessner2016}, NeuralTextures~\cite{Thies2019}, FaceSwap~\cite{FaceSwap}, and Deepfakes~\cite{Deepfakes}  are shown in Fig.~\ref{fig:examples_ff_Celeb-DF}.

\subsection{Implementation Details and Evaluation Settings} \label{subsec:imple}
In each image, we crop the face region and resize it to $256 \times 256$. In the neural cell search process, we set the batch size as $8$, the total epoch number as $300$. To optimize the architecture parameter $\alpha$, we use an Adam optimizer and initialize it with a learning rate of $0.02$. The learning rate is decayed by 10 at $60_{th}$ and $150_{th}$ epoch. To optimize the operation parameter $w$, we use a momentum SGD with a momentum value of $0.9$. We set the learning rate to $0.1$ and decrease it by a cosine scheduler. For both SGD and Adam optimizers, the weight decay number is set as $0.0005$.

After searching the neural cell, we use it to construct a deep architecture and finetune its parameters based on the target dataset. In this stage, we set the batch size as $96$, the learning rate as $0.5$, the epoch number as $300$. An Adam optimizer with a momentum value of $0.9$ and a weight decay of $0.0005$ are utilized. We decay the learning rate by 10 at the $80_{th}$ and $140_{th}$ epochs.

We test our searched architecture under two evaluation settings,~\textit{i.e.}, inner-dataset evaluation and cross-dataset evaluation. In the inner-dataset evaluation, the training set and testing set are manipulated by the same method, while in the cross-dataset evaluation, the training set and testing set are manipulated by two different methods. The cross-dataset evaluation scenario is utilized to investigate the robustness of the method under an unseen manipulation method. To make a fair comparison with prior works, we apply the Accuracy score (ACC) and Area Under the Receiver Operating Characteristic Curve (AUC) as our evaluation metrics.


\begin{table*}[h]\setlength{\tabcolsep}{11pt}
\centering
\caption{Performance comparisons on Deepfakes~\cite{Deepfakes}, Face2Face~\cite{MatthiasNiessner2016}, FaceSwap~\cite{FaceSwap}, and NeuralTextures~\cite{Thies2019}, and FF++~\cite{Thies2019} (High Quality).}
{
\begin{tabular}{lcccccccc}
\hline
\textbf{Method}  & \textbf{Venue} &\textbf{Input}& \textbf{Mask} & \textbf{Deepfakes} & \textbf{Face2Face} & \textbf{FaceSwap} & \textbf{NeuralTexture} & \textbf{FF++}\\   
\hline
Steg+SVM~\cite{fridrich2012rich}      &TIFS,2012 &RGB &N & 77.12\%  & 74.68\%  & 79.51\% & 76.94\% & 70.97\%\\
Bayar~\cite{bayar2016deep}  &IHMS,2016   &RGB  &N  & 90.18\%  & 94.93\% & 93.14\%  & 86.04\%  & 82.97\%\\
Cozzolino~\cite{cozzolino2017recasting}  &IHMS,2017  &RGB &N    & 81.78\%  & 85.32\%   & 85.69\%  &80.60\% & 78.45\%\\
Rahmouni~\cite{rahmouni2017distinguishing}   &WIFS,2017 &RGB &N  & 82.16\% & 93.48\%  & 92.51\% & 75.18\% & 79.08\%\\
MultiTask \cite{Nguyen2019_bats2019} &BATS,2019 &RGB &Y & 93.92\% &92.77\% &- & 88.05\% &-\\
LAE~\cite{Du2020}   &CIKM,2020 &RGB &Y &- & 92.14\%  &- &- &-\\
MseoNet~\cite{Hussain2021_wacv}  &WACV,2021 &RGB    &N  & 89.55\%  & 88.60\%  & 81.24\% & 76.62\% & 83.10\%\\
XceptionNet~\cite{Hussain2021_wacv}  &WACV,2021 &RGB &N & 97.49\% & 97.69\%  & 96.79\%  & 92.19\% & 92.39\%\\
SPSL~\cite{Liu2021_cvpr2021}  &CVPR,2021 &Freq &N & - & -  & -  & - & 91.50\%\\
\hline
ADD  &2021 &RGB &Y &97.45\% &98.33\% & 97.20\% & 90.84\% & 91.71\%\\ \hline
\end{tabular}
}
\label{tab:inncer_ffpp}
\end{table*}

As a common practice, we make comparisons with the following works:
\begin{itemize}
    \item Steg+SVM~\cite{fridrich2012rich} is based on handcrafted features and a linear support vector machine classifier.
    \item Cozzolino~\cite{cozzolino2017recasting} extends Steg+SVM~\cite{fridrich2012rich} by combining the hand-crafted features utilized in ~\cite{fridrich2012rich} with a convolutional neural network.
    \item Bayar~\cite{bayar2016deep} manually designs a convolutional neural network consisted of convolutional layers, max-pooling layers, and fully-connection layers.
    \item Rahmouni~\cite{rahmouni2017distinguishing} also proposes a convolutional neural architecture to extract features for deepfake detection.
    \item MseoNet~\cite{afchar2018mesonet} utilizes InceptionNet for detecting fake images. There are two inception modules in its architecture. The network is trained using a mean squared error between ground truth and prediction.
    \item XceptionNet~\cite{Hussain2021_wacv} uses an XcepiontNet which was manually designed for natural image classification. This architecture is constructed by inception-wise modules, in which depthwise separable convolution is utilized.
    \item LAE~\cite{Du2020} utilizes an encoder-decoder network to simultaneously predict real and fake labels as well as locate the manipulated regions. In LAE~\cite{Du2020}, the exact information of the manipulation method is required during training.
    \item SPSL~\cite{Liu2021_cvpr2021} manually designs a network to capture the forgery patterns from spatial image and phase spectrum.
    \item MultiTask \cite{Nguyen2019_bats2019} designs a convolutional neural network to simultaneously classify manipulated images and locate the manipulated regions in the corresponding images. Their designed network employs an encoder-decoder structure. In \cite{Nguyen2019_bats2019}, three tasks are learned simultaneously, which include: fake image detection, manipulated region segmentation, and image reconstruction. In MultiTask \cite{Nguyen2019_bats2019}, the exact information of the manipulation method is required during training.
    \item Context \cite{Nirkin2021} leverages the appearance discrepancy between the manipulated face region and its context to detect forgery face images. In Context \cite{Nirkin2021}, the exact information of the manipulation method is required for training its manually designed network.
\end{itemize}

We do not compare our work with Face X-ray \cite{li2020face_cvpr2020} since it needs external data collection and annotation for training their network.

\subsection{Comparison with previous methods}
\label{subsec:comparison}
\textbf{Inner-dataset evaluation}. The inner-dataset performance comparison on FF++ dataset is shown in Table~\ref{tab:inncer_ffpp}, which includes the performance comparison on: Deepfakes, Face2Face, FaceSwap, NeuralTexutre, and all manipulation methods as a whole (FF++). As shown in Table~\ref{tab:inncer_ffpp}, we can find that our method (ADD) outperforms previous non-deep learning methods ( Steg+SVM~\cite{fridrich2012rich} and Cozzolino~\cite{cozzolino2017recasting}) by a large margin ($10\% \sim 20\%$). Compared to Bayar~\cite{bayar2016deep}, Rahmouni~\cite{rahmouni2017distinguishing}, MseoNet~\cite{afchar2018mesonet}, our ADD still achieves a higher performance ($5\%\sim10\%$). LAE~\cite{Du2020} and MultiTask \cite{Nguyen2019_bats2019} are two methods trying to segment the manipulation regions based on the prior knowledge about the utilized manipulation method. Although ADD does not need the prior knowledge like LAE~\cite{Du2020} and MultiTask \cite{Nguyen2019_bats2019}, it still achieves a better performance. The superior performance demonstrates the efficacy of ADD. Among all the listed previous works, ADD is only outperformed by XceptionNet \cite{Hussain2021_wacv} on NeuralTexture and FF++ by a small margin; while on Deepfakes, Face2Face, and FaceSwap, ADD achieves competitive performance with XceptionNet. We assume that the performance difference between ADD and XceptionNet might come from the larger model capacity of XceptionNet. As shown in Table \ref{tab:model_size}, XceptionNet has more model parameters than ADD. 

In Table \ref{tab:model_size}, we report the performance comparison on Celeb-DF dataset as well as a model capacity comparison. Celeb-DF dataset provides a validation list. We utilize the samples not on the validation list to train four representative convolutional neural architectures, including ResNet-18 \cite{he2016deep}, ResNet-152 \cite{he2016deep}, VGG-16 \cite{simonyan2014very}, and XceptionNet. In Table \ref{tab:model_size}, we can find that our ADD achieves a competitive result comparing to ResNet-18 with a similar model capacity. Comparing to ResNet-152/VGG-16, our method still achieves a comparable performance with a smaller model capacity.

\textbf{Cross-dataset evaluation}. The performance comparison under the cross-dataset setting is shown in Table~\ref{tab:cross-dataset_face2face_faceswap} and Table \ref{tab:cross-dataset_ffpp_Celeb-DF}. Cross-dataset evaluation is more practical since in real scenarios, it is usually hard to know which manipulation method is applied on testing data. The evaluation results on cross-dataset settings can demonstrate the transferability of the method. 

To make a comparison with previous works~\cite{afchar2018mesonet, Rossler2019, Du2020}, we follow their experimental settings: (1) train the model on Face2Face and test it on FaceSwap (Table~\ref{tab:cross-dataset_face2face_faceswap}); (2) train the model on FF++ and test it on Celeb-DF (Table \ref{tab:cross-dataset_ffpp_Celeb-DF}). Not surprisingly, there is a drastic performance drop on all compared methods on the cross-dataset evaluation setting. As shown in both tables, our ADD outperforms previous works, such as~\cite{afchar2018mesonet, Rossler2019,Du2020}, by a large margin. The better performance demonstrating the superiority of our method might benefit from two aspects: on one hand, explicitly learn the potential manipulation regions make our network focus on important face regions to make correct prediction (as shown in Fig. \ref{fig:attention_add}); on the other hand, compared to works such as \cite{Du2020}, our method depends less on prior knowledge and mitigates the performance degradation introduced by the discrepancy of manipulated regions between different manipulation methods (as shown in Fig. \ref{fig:manual_auto}).

In our method, generating ground truth for potential manipulation regions has its own advantages. In previous works, drawing ground truth for manipulation regions is time-consuming and prior knowledge cost; in contrast,  our method generates the ground truth for potential manipulation regions in a highly efficient manner (less than one second) with few dependence on the beforementioned prior knowledge.


\begin{table}[!h]\setlength{\tabcolsep}{9pt}
\centering
\caption{Cross-dataset evaluation results. Train on Face2Face and test on FaceSwap.}
{
\begin{tabular}{lcccc}
\hline
\textbf{Method}  & \textbf{Venue} & \textbf{Mask} & \textbf{Input} & \textbf{ACC} \\
\hline
MseoNet~\cite{afchar2018mesonet}  &WIFS,2018   &N & RGB     & 47.32\%   \\
XceptionNet~\cite{Rossler2019}  &ICCV,2019  &N & RGB  & 49.94\%   \\
LAE~\cite{Du2020}   &CIKM,2020  &Y & RGB  & 63.15\%  \\
\hline
ADD &2021  &Y & RGB &67.02\% \\ \hline
\end{tabular}
}
\label{tab:cross-dataset_face2face_faceswap}
\end{table}

\subsection{Ablation Study} \label{subsec:ablation}
\begin{figure}[htbp]
  \centering
  \includegraphics[width=1.0\linewidth]{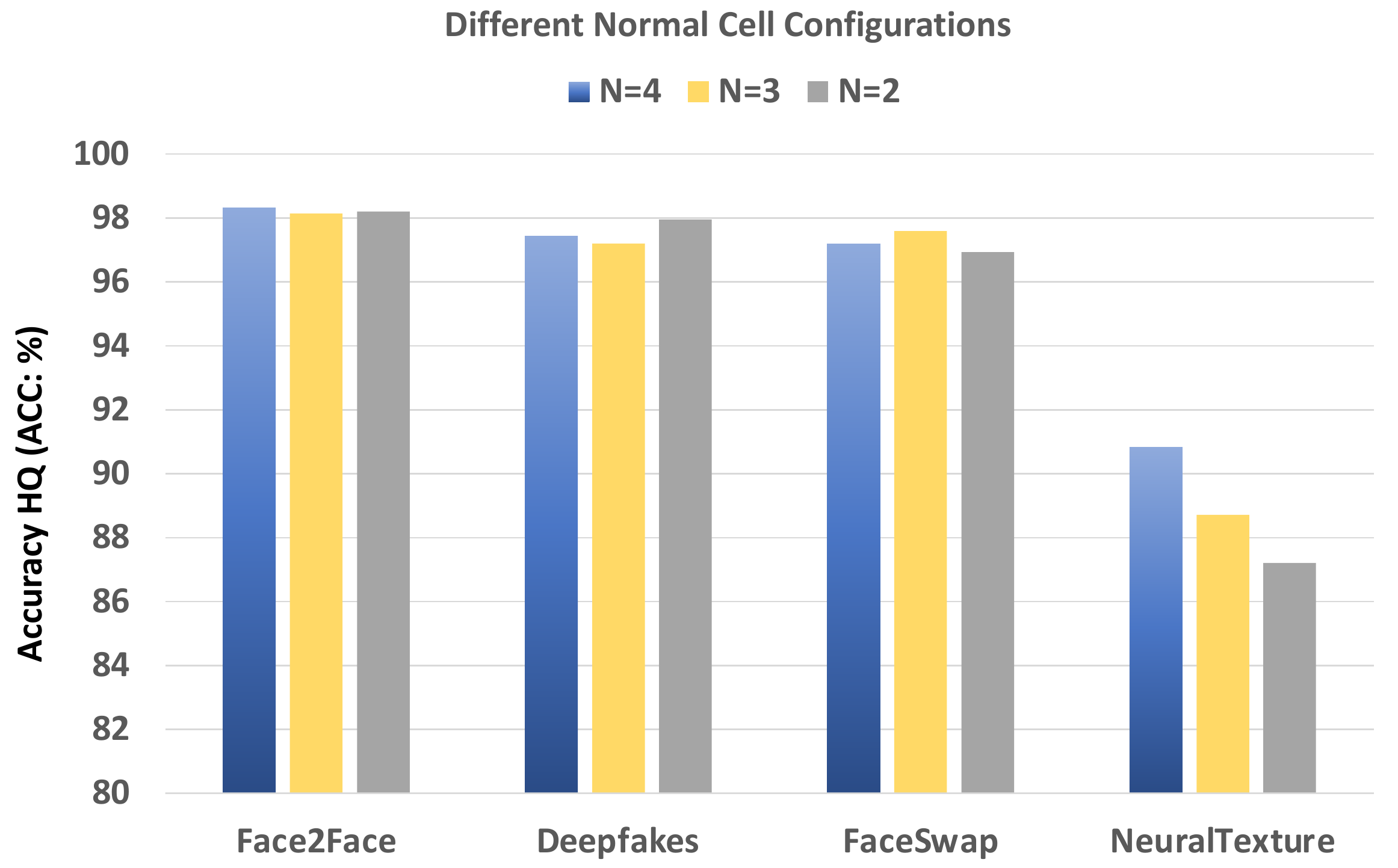}  
  \caption{ {Performance comparison with different normal cell configurations on Face2Face~\cite{MatthiasNiessner2016}, Deepfakes~\cite{Deepfakes}, FaceSwap~\cite{FaceSwap}, and NeuralTextures~\cite{Thies2019}. }}
  \label{fig:differentmacro}
\end{figure}

\textbf{Different normal cell configurations} By hierarchically stacking the normal cell \textit{different} times, we construct deep models with different capacities. We make a performance comparison between them on four manipulation methods in FF++ dataset. The performance comparison is shown in Fig. \ref{fig:differentmacro}. From the figure, we can find that the network with a larger capacity (a larger stacking number $N$) achieves a better performance, which matches the previous findings in Table \ref{tab:model_size}. Unless stated, the stacked number utilized in our work is set as $4$.



\begin{figure}[htbp]
  \centering
  \includegraphics[width=1.0\linewidth]{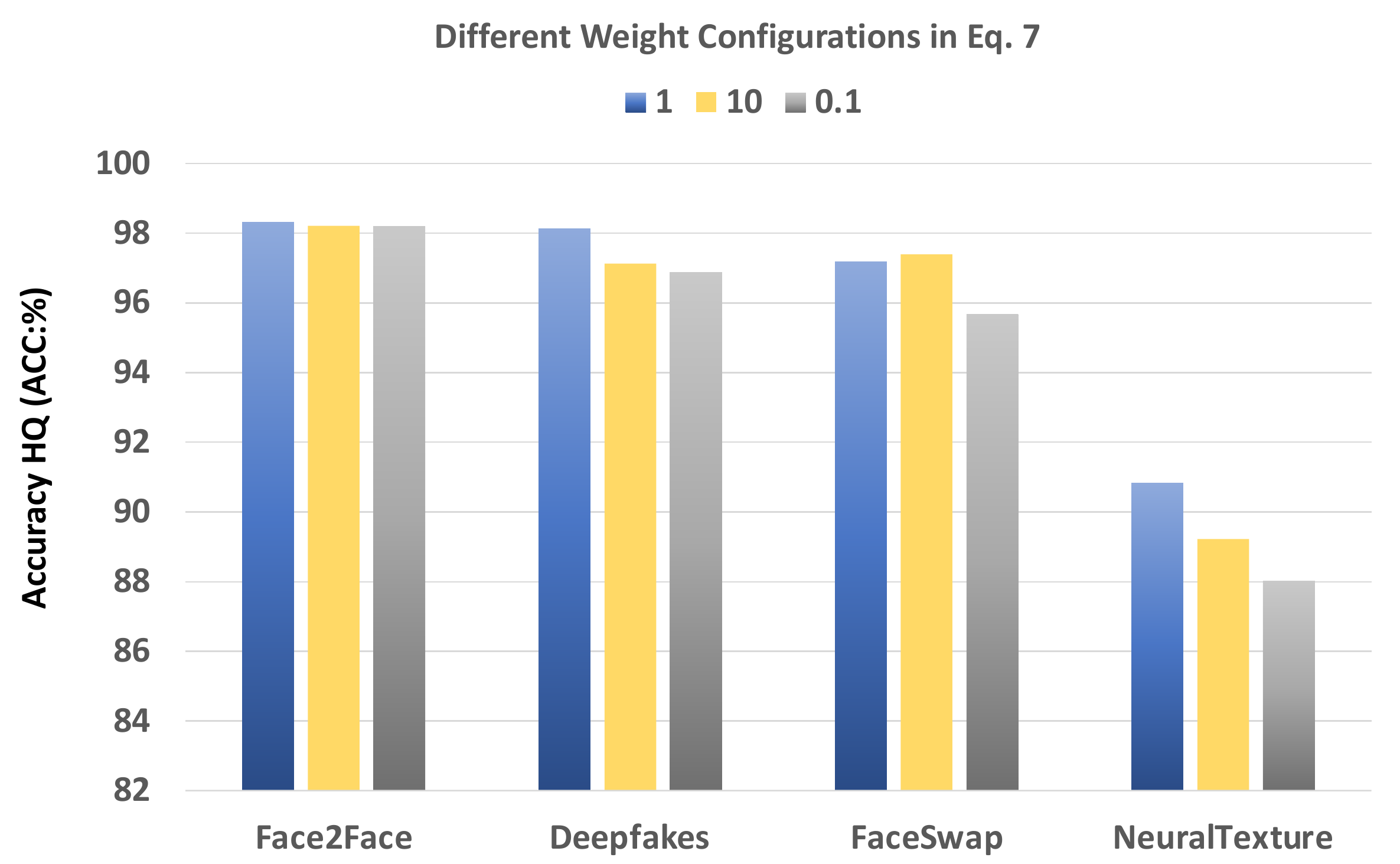}  
  \caption{ {Performance comparison with different loss weight configurations on Face2Face~\cite{MatthiasNiessner2016}, Deepfakes~\cite{Deepfakes}, FaceSwap~\cite{FaceSwap}, and NeuralTextures~\cite{Thies2019}. }}
  \label{fig:differentweights}
\end{figure}

\textbf{Different weight configurations in Eq. \ref{eq:l_overall}} By setting $\alpha$ in Eq. \ref{eq:l_overall} as different non-zero values, \textit{i.e.}, $1$, $10$, $0.1$, we conduct a performance comparison about the loss term contributions. The comparison result is shown in Fig. \ref{fig:differentweights}. From the figure, we can find that  $\alpha$=1 achieves better performance on Face2Face and Deepfakes. On FaceSwap, $\alpha$=1 is outperformed by $\alpha$=10 a bit. On NeuralTexture, $\alpha$=1 outperforms other two by a large margin. Based on the observation, we set $\alpha$ as 1 in our experiment.


\textbf{Efficacy of potential manipulation region estimation} In Eq. \ref{eq:l_overall}, by setting $\alpha$ as zero (w/o $\mathcal{L}_{MSE}$, the single task setting) or not (w $\mathcal{L}_{MSE}$, the multi-task setting), we conduct an experiment on cross-dataset evaluations to demonstrate the efficacy of our learning potential manipulation region strategy. The performance comparison is shown in Table \ref{tab:with_out_mask}. From the table, we can find that the performance with $\mathcal{L}_{MSE}$ is much higher than the performance without $\mathcal{L}_{MSE}$. This indicates the efficacy and necessity of learning the potential manipulation regions in ADD.

\begin{table}[H]\setlength{\tabcolsep}{9pt}
\centering
\caption{Cross-dataset evaluation results. Train on FF++ training set and test on Celeb-DF.}
{
\begin{tabular}{lcccc}
\hline
\textbf{Method}  & \textbf{Venue}  & \textbf{Mask}  & \textbf{Input} & \textbf{AUC} \\
\hline
MseoNet~\cite{afchar2018mesonet}  &WIFS,2018 &N  &RGB    &54.80\%  \\
Capsule~\cite{Nguyen2019}  &ICASSP,2019  &N   &RGB  & 57.50\%\\
MultiTask \cite{Nguyen2019_bats2019} &BATS,2019 &Y  &RGB  & 54.30\%\\
VA-MLP \cite{8638330_wacv2019} &WACV,2019 &N &RGB  & 55.00\%\\
SMIL \cite{Li2020_mm2020} &MM,2020 &N  &RGB  & 56.30\%\\
F3-Net \cite{Qian2020_eccv2020} &ECCV,2020 &N  &Freq  & 65.17\%\\
XceptionNet~\cite{Liu2021_cvpr2021}  &CVPR,2021   &N  &RGB & 65.50\%\\
Context~\cite{Nirkin2021}  &TPAMI,2021   &Y  &RGB & 66.00\%\\
\hline
ADD &2021   &Y   &RGB & 66.48\%\\ \hline
\end{tabular}
}
\label{tab:cross-dataset_ffpp_Celeb-DF}
\end{table}


\begin{table}[H]\setlength{\tabcolsep}{4.2pt}
\centering
\caption{Accuracy and model size comparison on Celeb-DF.}
{
\begin{tabular}{lccccc}
\hline
\textbf{Backbone}  & \textbf{ResNet-18}  & \textbf{ResNet152} & \textbf{VGG-16} & \textbf{XceptionNet} & \textbf{ADD}\\
\hline
ACC &96.76\%    & 98.30\%  & 97.88\% &96.32\%   & 97.57\% \\
Params (M) & 11.6 &60.3 & 138  &23 &5.6 \\
\hline

\end{tabular}
}
\label{tab:model_size}
\end{table}

\begin{table}[H]\setlength{\tabcolsep}{15pt}
\centering
\caption{Performance compassion between w and w/o $\mathcal{L}_{MSE}$.}
{
\begin{tabular}{lccc}
\hline
\textbf{Method}  & \textbf{Source} & \textbf{Target} & \textbf{ACC}  \\
\hline
w/o $\mathcal{L}_{MSE}$    &FF++ & Celeb-DF  & 64.81\%     \\
w $\mathcal{L}_{MSE}$   &FF++ & Celeb-DF & 69.19\%  \\ \midrule
w/o $\mathcal{L}_{MSE}$    &Face2Face & FaceSwap  & 57.22\%   \\
w $\mathcal{L}_{MSE}$     &Face2Face & FaceSwap & 67.02\%    \\
\hline
\end{tabular}
}
\label{tab:with_out_mask}
\end{table}



\begin{figure}[htbp]
  \centering
  \includegraphics[width=1.0\linewidth]{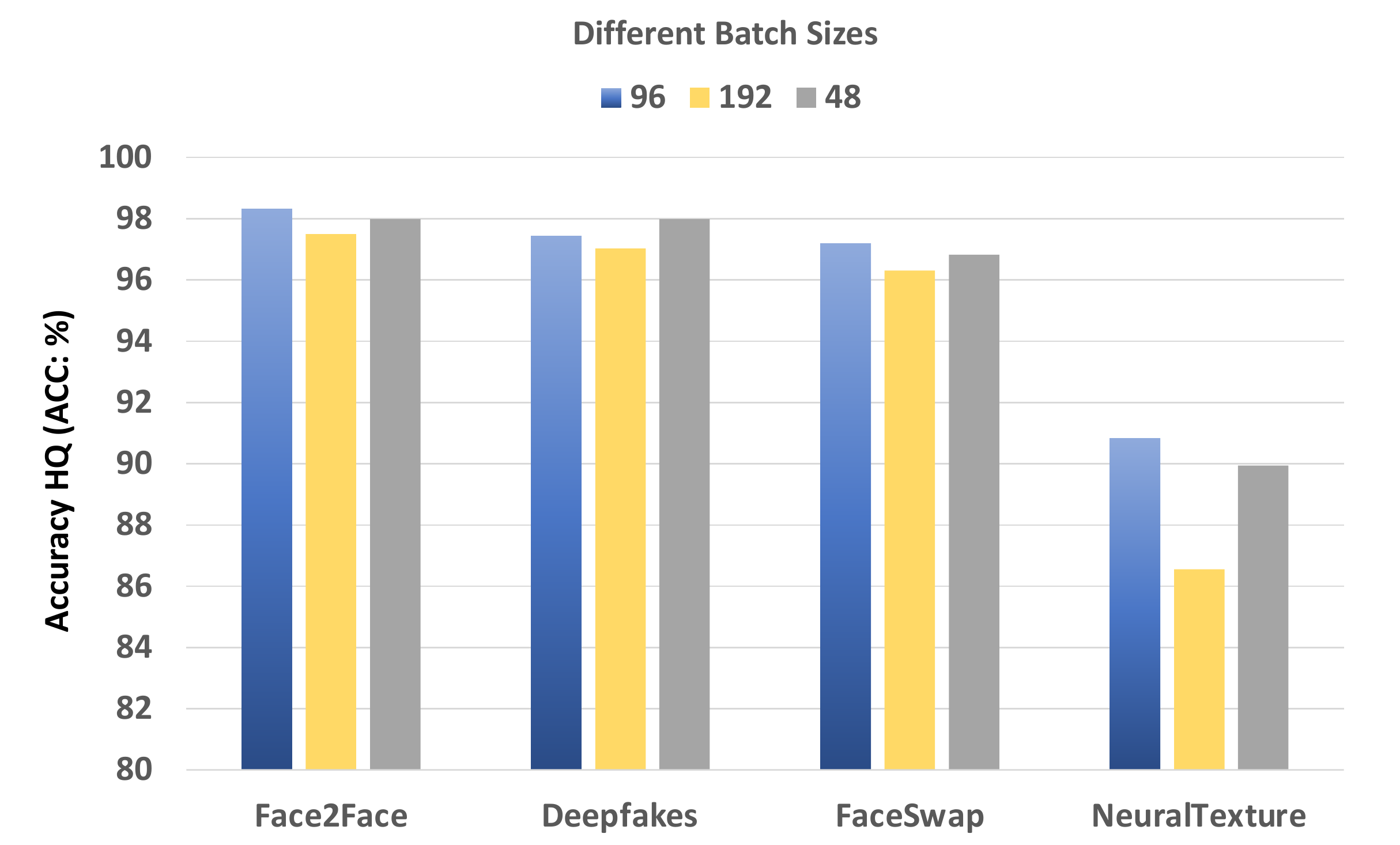}  
  \caption{ {Performance comparison with different batch sizes on Face2Face~\cite{MatthiasNiessner2016}, Deepfakes~\cite{Deepfakes}, FaceSwap~\cite{FaceSwap}, and NeuralTextures~\cite{Thies2019}. }}
  \label{fig:differentbatchsizes}
\end{figure}

\textbf{Comparison of different batch sizes} We test the performance of our searched cell under different training batch sizes. Specifically, by fixing the learning rate as $0.5$, stacking number as $4$, we set the batch size as $96$, $192$, and $48$, and test them on four manipulation methods in FF++ dataset. The performance comparison is shown in Fig. \ref{fig:differentbatchsizes}. From the figure, we can find that the performance with a batch size of $96$ outperforms that with a batch size of $192$. Our speculation is that using a too large batch size might cause a poor generalization performance, which is known as the ``generalization gap" phenomenon and discussed in previous works such as \cite{Houlden2017_nips2017}.




\begin{figure}[htbp]
  \centering
  \includegraphics[width=1.0\linewidth]{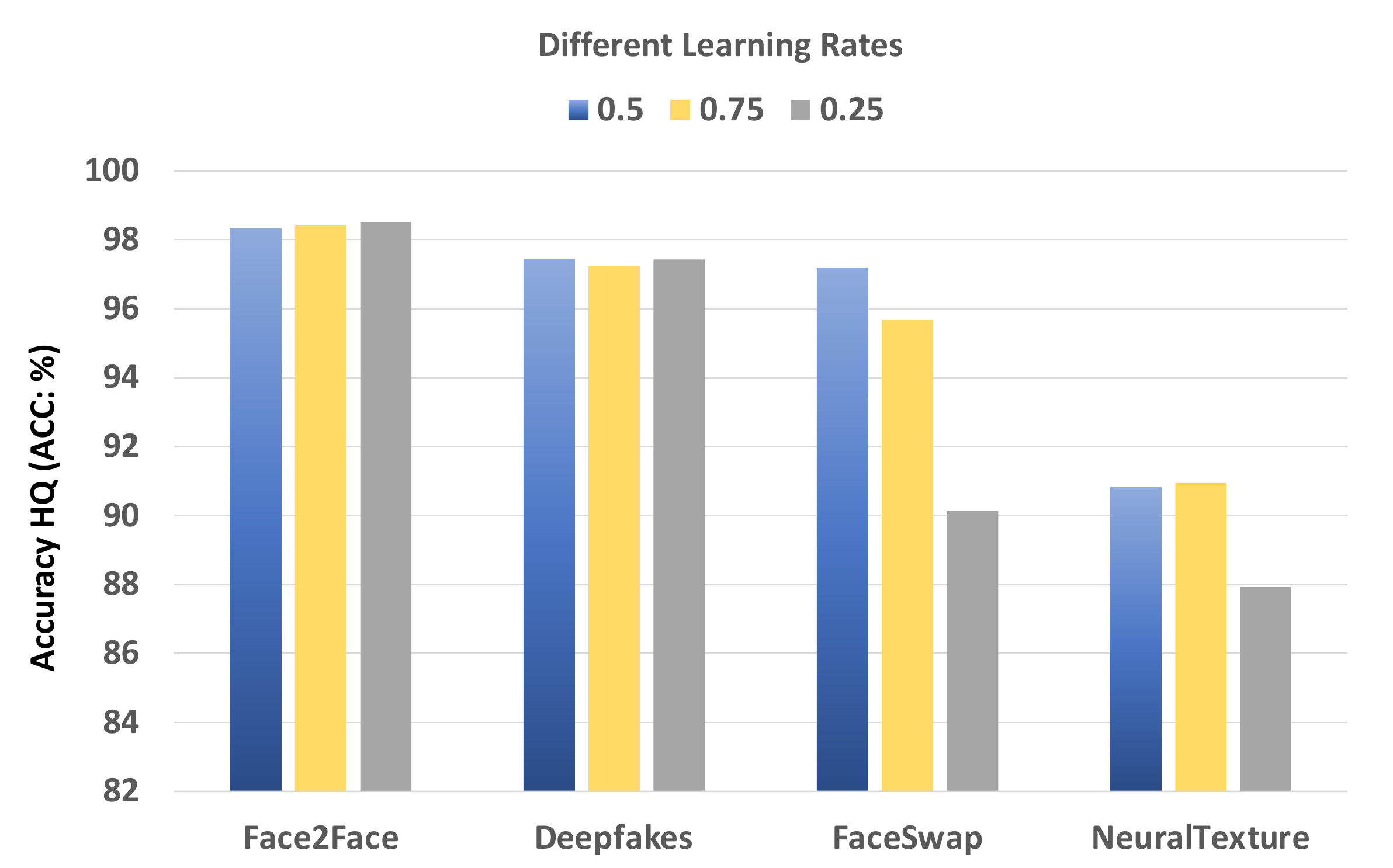}  
  \caption{ {Performance comparison with different learning rates on Face2Face~\cite{MatthiasNiessner2016}, Deepfakes~\cite{Deepfakes}, FaceSwap~\cite{FaceSwap}, and NeuralTextures~\cite{Thies2019}. }}
  \label{fig:differentlearningrates}
\end{figure}

\textbf{Comparison of different learning rates} We conduct a performance comparison between different learning rates. Specifically, fixing the batch size as $96$, stacking number as $4$, we make comparisons between different learning rates of  $0.75$ and $0.5$.  We test the two different learning rates and show the performance comparisons in Fig. \ref{fig:differentlearningrates}. From Fig. \ref{fig:differentlearningrates}, we can find that the performance with a learning rate of $0.5$ outperforms a bit than the performance with a learning rate of $0.75$.


\textbf{Robustness to Low-Quality Input} We conduct an experiment on low-quality images. The low quality means videos are generated in lower quality levels, specifically, with quantization of $40$ \cite{Rossler2019}. By setting the batch size as $96$, stacking number as $4$, learning rate as $0.5$, we conduct an experiment on low-quality Deepfakes~\cite{Deepfakes} and show the performance comparisons in Table \ref{tab:lq_deepfakes}. From the table, we can find that although the performance of our method on low-quality input is lower than that on high-quality input, it still outperforms previous works. Specifically, on Deepfakes~\cite{Deepfakes} in LQ format, the performance of our method is quite near to F3-Net \cite{Qian2020_eccv2020} which mines forgery patterns in a frequency domain. As illustrated in ``Discussion" subsection, our method is complementary to \cite{Qian2020_eccv2020}, and we will conduct an exploration about how to apply ADD on the frequency domain in future investigations. Based on the results shown in Table \ref{tab:lq_deepfakes}, we believe that our method is robust to the low-quality data to some extent.


\begin{table}[!h]\setlength{\tabcolsep}{12pt}
\centering
\caption{Performance comparisons on low quality Deepfakes.}
{
\begin{tabular}{lccc}
\hline
\textbf{Method}  & \textbf{Venue}  &\textbf{Input} & \textbf{ACC} \\
\hline
Steg+SVM~\cite{fridrich2012rich}      &TIFS,2012 &RGB & 67.00\%  \\
Bayar~\cite{bayar2016deep}  &IHMS,2016    &RGB   &87.00\%   \\
Cozzolino~\cite{cozzolino2017recasting}  &IHMS,2017  &RGB     & 75.00\%   \\
Rahmouni~\cite{rahmouni2017distinguishing}   &WIFS,2017  &RGB  & 80.00\%   \\
MseoNet~\cite{afchar2018mesonet}  &WIFS,2018    &RGB   & 90.00\%   \\
XceptionNet~\cite{Rossler2019}  &ICCV,2019 &RGB & 96.01\%   \\
F3-Net~\cite{Qian2020_eccv2020}   &ECCV,2020 &Freq &97.97\%  \\
\hline
ADD &2021 &RGB &97.03\% \\ \hline
\end{tabular}
}
\label{tab:lq_deepfakes}
\end{table}


\subsection{Visualization} \label{subsec:visualization}
\begin{figure}[htbp]
  \centering
  \includegraphics[width=1\linewidth]{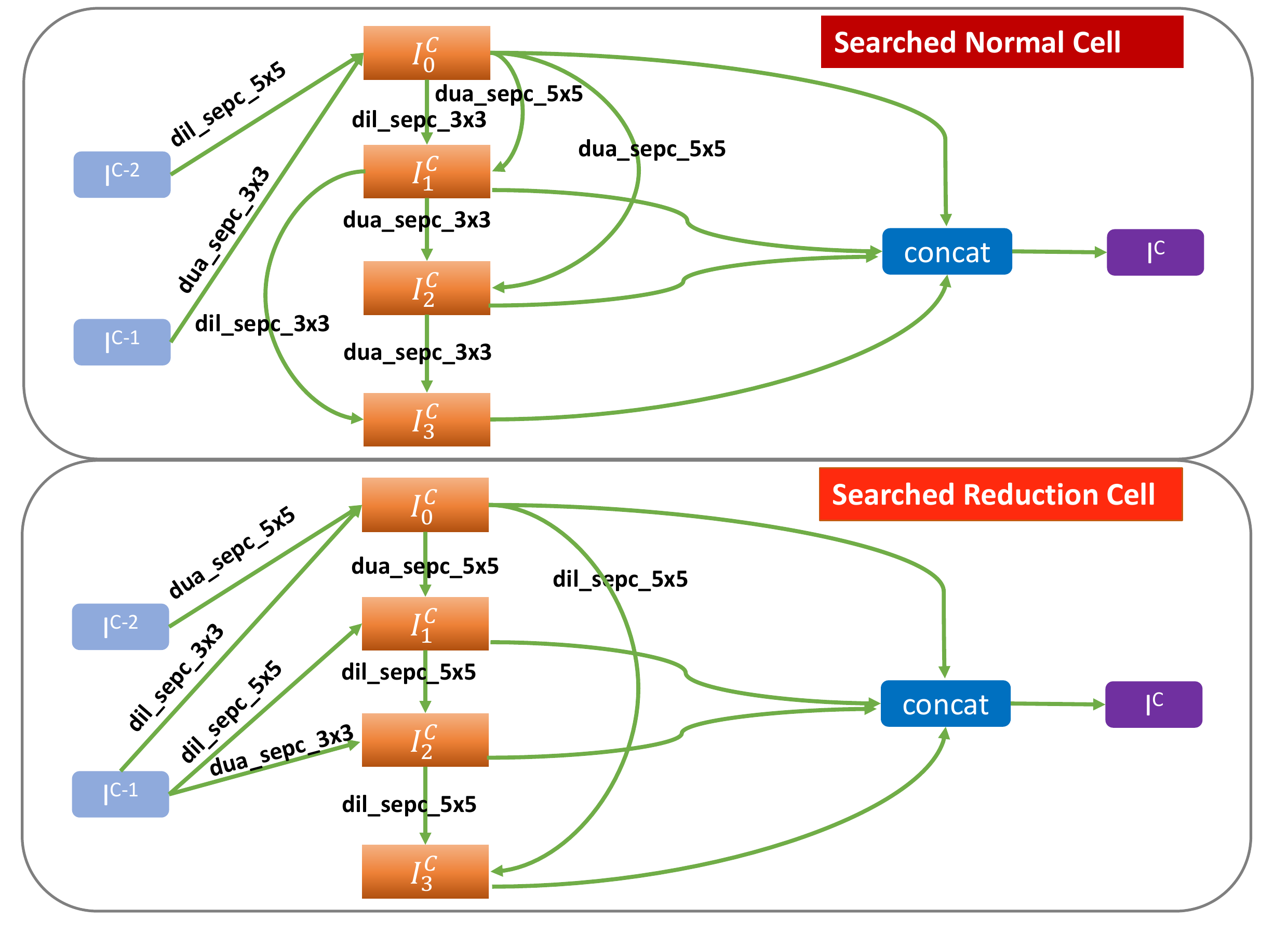}  
  \caption{ {The visualization of the normal cell and the reduction cell searched by our methods. The selected operation definitions can be found in Sec. \ref{subsec:sdd}. }}
  \label{fig:searched_cell}
\end{figure}

\textbf{Searched Cells} To comprehend the neural cell architecture searched by our method, we display our searched results, \textit{i.e.}, the normal cell and reduction cell, in Fig. \ref{fig:searched_cell}. As shown in the figure, the automatically searched cells are quite complex and different from any existing neural architectures. It is difficult to design those cells for human engineers. An interesting finding is that both normal cell and reduction cell searched by our method extensively utilize separable convolution and dilated convolution. This matches the design principle manually proposed in XceptionNet \cite{Rossler2019}. However, our method discovers this principle in an automated manner. Another discovery is that the operations with a large kernel size, \textit{i.e.}, $7 \times 7$, are not selected. There are two possible reasons for this: (1) deepfake detection is different from dense prediction tasks such as semantic segmentation \cite{Peng_2017_CVPR}, therefore a large size kernel is not indispensable for constructing a network for deepfake detection; (2) as pointed in \cite{simonyan2014very}, a $7 \times 7$ convolution kernel can be composed through a set of  $3 \times 3$ convolution filters, while the latter manner is preferred due to its less parameter numbers and computation cost.

\begin{figure}[htbp]
  \centering
  \includegraphics[width=1\linewidth]{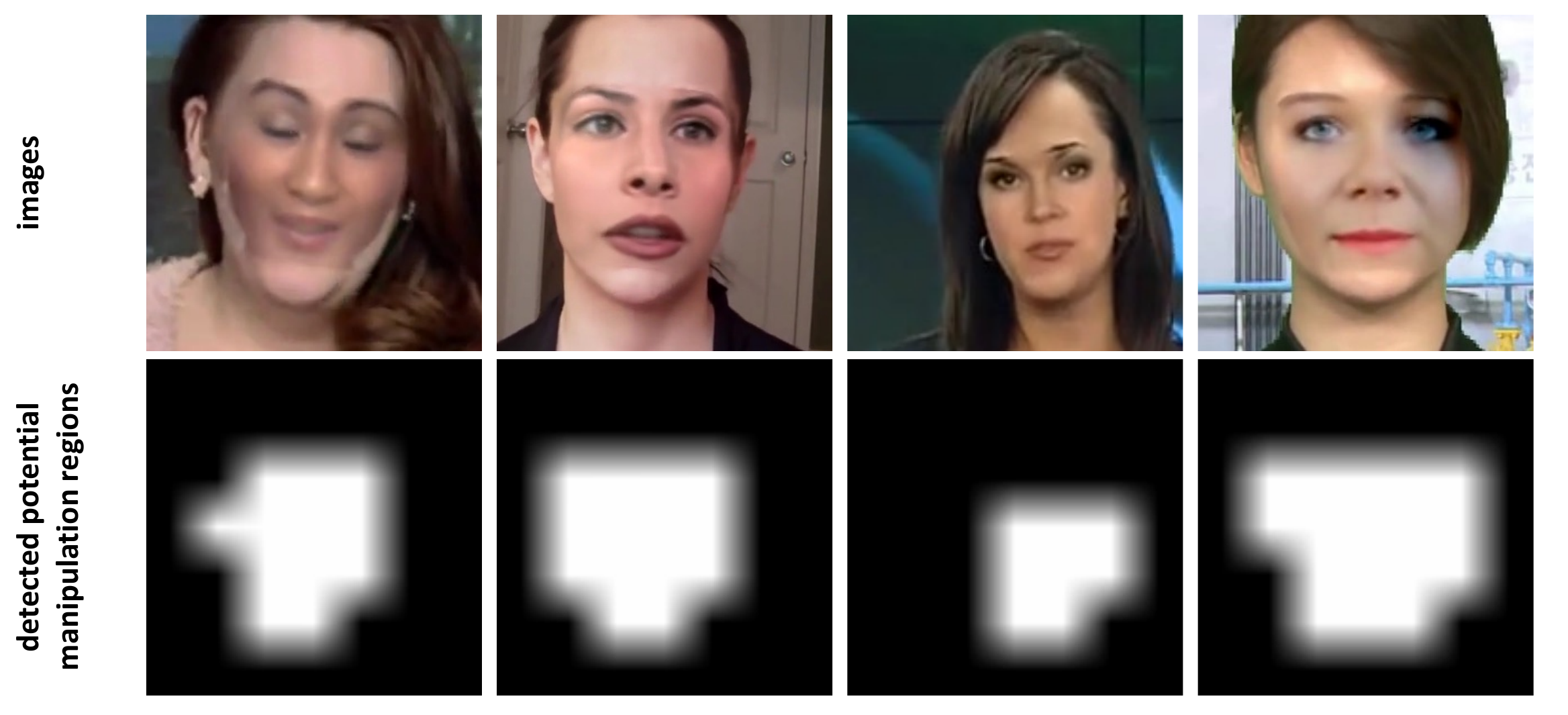}  
  \caption{ {Visualization of predicted potential manipulation regions by ADD. For training our method, ADD does not require the prior knowledge such as what manipulation method is applied.}}
  \label{fig:attention_add}
\end{figure}

\textbf{Visualization in terms of potential manipulation region localization}
Our ADD conducts deepfake detection based on the focused features inside the localized potential manipulation regions. In Fig. \ref{fig:attention_add}, we show that our method can correctly localize the potential manipulation region in each given sample, given no clue about which manipulation method is applied. It can be seen from the figure that the predicted potential manipulation region in each sample matches the position, shape, and scale of the corresponding images.

\subsection{Discussion} \label{subsec:discussions}
We are aware that there are three possible research directions for our ADD. 1) In this submission, our method conducts an architecture search based on RGB images and spatial domains only. With the development of forgery image generation methods \cite{Li_2020_CVPR,Wang2021}, the local forgery patterns in the spatial domain become too subtle to detect. Recent works, such as \cite{Qian2020_eccv2020, Shen2021_aaai2021}, point out the frequency domain provides complementary information for deepfake detection. They propose to mine discriminative forgery patterns from frequency domains by translating original RGB images to frequency images. Our proposed ADD can be applied to frequency domains to detect forgery clues with few modifications. More than that, with an adaptive fusion strategy (for example, cross-modal transformer module \cite{zhang2021prouigan_arxiv2021}), it is expected that a RGB ADD (working on spatial domains) and a frequency ADD (working on frequency domains) can collaborate with each other to further boost the deepfake detection accuracy; 2) besides the potential manipulation region localization, there are other related facial analysis tasks that can be utilized to provide an additional signal for network training. For example, Mazaheri \textit{et al.} \cite{Mazaheri2021} propose to integrate a facial expression recognition module for improving deepfake detection accuracy; 3) it is worth exploring how to apply our method to search a recurrent neural network for detecting forgery patterns in sequential data \cite{qi2020deeprhythm_acmmm_2020,wang2020deepsonar_acmmm2020}. Future investigation in those directions will be placed.


\subsection{Limitations} \label{subsec:limitations}
Although our method demonstrates its effectiveness on deepfake detection experimentally, it still has some limitations: 1) the potential manipulation region learning strategy might not work well if the fake image is entirely synthetic. This is because when the forgery sample is entirely synthetic, there is no manipulation region anymore. In such a case, ADD w/o the potential manipulation region localization mechanism should be utilized. 2) We notice that our method still suffers from a performance drop when encountering low-quality images. We would explore novel strategies to improve the robustness of our method when dealing with low-quality images.

\section{Conclusion}
In this article, we propose to utilize automated machine learning to automatically construct a neural architecture for deepfake detection. To the best of our knowledge, this is the first time to apply AutoML to this research problem. To improve the generalizability of our method, we integrate a simple yet efficient strategy in our network learning process, making it estimate the potential manipulation regions as well as predict the real or fake labels. Compared to previous works, our method depends less on prior knowledge, \textit{e.g.}, no need to know which manipulation methods is utilized or whether it is utilized. Experimental results on two benchmark datasets demonstrate the efficacy of our proposed method, especially when testing data and training data are manipulated by different methods. In the future, we plan to explore more advanced search methods and search spaces to further improve the prediction ability of our method.



%





\ifCLASSOPTIONcaptionsoff
  \newpage
\fi

\bibliographystyle{IEEEtran}
\bibliography{autofake}

\begin{thebibliography}{10}
\providecommand{\url}[1]{#1}
\csname url@samestyle\endcsname
\providecommand{\newblock}{\relax}
\providecommand{\bibinfo}[2]{#2}
\providecommand{\BIBentrySTDinterwordspacing}{\spaceskip=0pt\relax}
\providecommand{\BIBentryALTinterwordstretchfactor}{4}
\providecommand{\BIBentryALTinterwordspacing}{\spaceskip=\fontdimen2\font plus
\BIBentryALTinterwordstretchfactor\fontdimen3\font minus
  \fontdimen4\font\relax}
\providecommand{\BIBforeignlanguage}[2]{{%
\expandafter\ifx\csname l@#1\endcsname\relax
\typeout{** WARNING: IEEEtran.bst: No hyphenation pattern has been}%
\typeout{** loaded for the language `#1'. Using the pattern for}%
\typeout{** the default language instead.}%
\else
\language=\csname l@#1\endcsname
\fi
#2}}
\providecommand{\BIBdecl}{\relax}
\BIBdecl

\bibitem{abdal2019image2stylegan}
R.~Abdal, Y.~Qin, and P.~Wonka, ``Image2stylegan: How to embed images into the
  stylegan latent space?'' in \emph{ICCV}, 2019.

\bibitem{karras2020analyzing}
T.~Karras, S.~Laine, M.~Aittala, J.~Hellsten, J.~Lehtinen, and T.~Aila,
  ``Analyzing and improving the image quality of stylegan,'' in \emph{CVPR},
  2020.

\bibitem{wu2019relgan}
P.-W. Wu, Y.-J. Lin, C.-H. Chang, E.~Y. Chang, and S.-W. Liao, ``Relgan:
  Multi-domain image-to-image translation via relative attributes,'' in
  \emph{ICCV}, 2019.

\bibitem{DBLP:journals/corr/abs-2103-05193}
Y.~Shi, X.~Zhou, P.~Liu, and I.~W. Tsang, ``Generative transition mechanism to
  image-to-image translation via encoded transformation,'' \emph{CoRR}, 2021.

\bibitem{9110728}
B.~Hu, Z.~Zheng, P.~Liu, W.~Yang, and M.~Ren, ``Unsupervised eyeglasses removal
  in the wild,'' \emph{IEEE Transactions on Cybernetics}, 2020.

\bibitem{9318504}
Y.~Zhang, I.~W. Tsang, J.~Li, P.~Liu, X.~Lu, and X.~Yu, ``Face hallucination
  with finishing touches,'' \emph{IEEE Transactions on Image Processing}, 2021.

\bibitem{7086315_tifs2015}
C.-M. Pun, X.-C. Yuan, and X.-L. Bi, ``Image forgery detection using adaptive
  oversegmentation and feature point matching,'' \emph{IEEE Transactions on
  Information Forensics and Security}, 2015.

\bibitem{7154457_tifs2015}
D.~Cozzolino, G.~Poggi, and L.~Verdoliva, ``Efficient dense-field copy–move
  forgery detection,'' \emph{IEEE Transactions on Information Forensics and
  Security}, 2015.

\bibitem{6987281_tifs2015}
J.~Li, X.~Li, B.~Yang, and X.~Sun, ``Segmentation-based image copy-move forgery
  detection scheme,'' \emph{IEEE Transactions on Information Forensics and
  Security}, 2015.

\bibitem{9298826_tifs2021}
C.-Z. Yang, J.~Ma, S.~Wang, and A.~W.-C. Liew, ``Preventing deepfake attacks on
  speaker authentication by dynamic lip movement analysis,'' \emph{IEEE
  Transactions on Information Forensics and Security}, 2021.

\bibitem{9505637_tifs2021}
J.~Yang, A.~Li, S.~Xiao, W.~Lu, and X.~Gao, ``Mtd-net: Learning to detect
  deepfakes images by multi-scale texture difference,'' \emph{IEEE Transactions
  on Information Forensics and Security}, 2021.

\bibitem{8253869}
S.~Wang, B.~Pan, H.~Chen, and Q.~Ji, ``Thermal augmented expression
  recognition,'' \emph{IEEE Transactions on Cybernetics}, 2018.

\bibitem{9376703}
Y.~Yang, Y.~Hu, X.~Zhang, and S.~Wang, ``Two-stage selective ensemble of cnn
  via deep tree training for medical image classification,'' \emph{IEEE
  Transactions on Cybernetics}, 2021.

\bibitem{9376704_heheefan_2021}
H.~Fan, P.~Liu, M.~Xu, and Y.~Yang, ``Unsupervised visual representation
  learning via dual-level progressive similar instance selection,'' \emph{IEEE
  Transactions on Cybernetics}, 2021.

\bibitem{8933048_tifs2020}
G.~Hu, Y.~Xiao, Z.~Cao, L.~Meng, Z.~Fang, J.~T. Zhou, and J.~Yuan, ``Towards
  real-time eyeblink detection in the wild: Dataset, theory and practices,''
  \emph{IEEE Transactions on Information Forensics and Security}, 2020.

\bibitem{9346018}
J.~Zhou, L.~Zhang, D.~Jiawei, X.~Peng, Z.~Fang, Z.~Xiao, and H.~Zhu,
  ``Locality-aware crowd counting,'' \emph{IEEE Transactions on Pattern
  Analysis and Machine Intelligence}, 2021.

\bibitem{luo2020every_pr2020}
Y.~Luo, R.~Ji, T.~Guan, J.~Yu, P.~Liu, and Y.~Yang, ``Every node counts:
  Self-ensembling graph convolutional networks for semi-supervised learning,''
  \emph{Pattern Recognition}, 2020.

\bibitem{Peng2021Joint}
P.~Hu, X.~Peng, H.~Zhu, J.~Lin, L.~Zhen, and D.~Peng, ``Joint versus
  independent multiview hashing for cross-view retrieval,'' \emph{IEEE
  Transactions on Cybernetics}, 2020.

\bibitem{Miao_2019_ICCV}
J.~Miao, Y.~Wu, P.~Liu, Y.~Ding, and Y.~Yang, ``Pose-guided feature alignment
  for occluded person re-identification,'' in \emph{ICCV}, 2019.

\bibitem{Zheng2021}
Z.~Zheng and Y.~Yang, ``{Rectifying Pseudo Label Learning via Uncertainty
  Estimation for Domain Adaptive Semantic Segmentation},'' \emph{International
  Journal of Computer Vision}, 2021.

\bibitem{8485427}
Z.~Zhong, L.~Zheng, Z.~Zheng, S.~Li, and Y.~Yang, ``Camstyle: A novel data
  augmentation method for person re-identification,'' \emph{IEEE Transactions
  on Image Processing}, 2019.

\bibitem{9068282_tcsvt2021}
P.~Li, P.~Pan, P.~Liu, M.~Xu, and Y.~Yang, ``Hierarchical temporal modeling
  with mutual distance matching for video based person re-identification,''
  \emph{IEEE Transactions on Circuits and Systems for Video Technology}, 2021.

\bibitem{9108530}
X.~Zhang, Y.~Wei, Y.~Yang, and T.~S. Huang, ``Sg-one: Similarity guidance
  network for one-shot semantic segmentation,'' \emph{IEEE Transactions on
  Cybernetics}, 2020.

\bibitem{luo2018macro}
Y.~Luo, Z.~Zheng, L.~Zheng, T.~Guan, J.~Yu, and Y.~Yang, ``Macro-micro
  adversarial network for human parsing,'' in \emph{ECCV}, 2018.

\bibitem{luo2019significance}
Y.~Luo, P.~Liu, T.~Guan, J.~Yu, and Y.~Yang, ``Significance-aware information
  bottleneck for domain adaptive semantic segmentation,'' in \emph{ICCV}, 2019.

\bibitem{luo2020ASM}
------, ``Adversarial style mining for one-shot unsupervised domain
  adaptation,'' in \emph{NeurIPS}, 2020.

\bibitem{pan2020adversarial}
P.~Pan, P.~Liu, Y.~Yan, T.~Yang, and Y.~Yang, ``Adversarial localized energy
  network for structured prediction,'' in \emph{AAAI}, 2020.

\bibitem{he2016deep}
K.~He, X.~Zhang, S.~Ren, and J.~Sun, ``Deep residual learning for image
  recognition,'' in \emph{CVPR}, 2016.

\bibitem{chollet2017xception}
F.~Chollet, ``Xception: Deep learning with depthwise separable convolutions,''
  in \emph{CVPR}, 2017.

\bibitem{Du2020}
M.~Du, S.~Pentyala, Y.~Li, and X.~Hu, ``{Towards Generalizable Forgery
  Detection with Locality-aware AutoEncoder},'' in \emph{CIKM}, 2020.

\bibitem{liu2020global}
Z.~Liu, X.~Qi, and P.~H. Torr, ``Global texture enhancement for fake face
  detection in the wild,'' in \emph{CVPR}, 2020.

\bibitem{Zhou2021_cvpr2021}
T.~Zhou, W.~Wang, Z.~Liang, and J.~Shen, ``{Face Forensics in the Wild},'' in
  \emph{CVPR}, 2021.

\bibitem{Dong_2019_CVPR}
X.~Dong and Y.~Yang, ``Searching for a robust neural architecture in four gpu
  hours,'' in \emph{CVPR}, 2019.

\bibitem{liu2018darts}
H.~Liu, K.~Simonyan, and Y.~Yang, ``Darts: Differentiable architecture
  search,'' in \emph{ICLR}, 2019.

\bibitem{Mazaheri2021}
G.~Mazaheri and A.~K. Roy-Chowdhury, ``{Detection and Localization of Facial
  Expression Manipulations},'' \emph{arXiv}, 2021.

\bibitem{Rossler2019}
A.~Rossler, D.~Cozzolino, L.~Verdoliva, C.~Riess, J.~Thies, and M.~Niessner,
  ``{FaceForensics++: Learning to detect manipulated facial images},'' in
  \emph{ICCV}, 2019.

\bibitem{Li2020_cvpr2020}
Y.~Li, X.~Yang, P.~Sun, H.~Qi, and S.~Lyu, ``{Celeb-DF : A Large-scale
  Challenging Dataset for DeepFake Forensics},'' in \emph{CVPR}, 2020.

\bibitem{goodfellow2014generative_nips}
I.~J. Goodfellow, J.~Pouget-Abadie, M.~Mirza, B.~Xu, D.~Warde-Farley, S.~Ozair,
  A.~Courville, and Y.~Bengio, ``Generative adversarial networks,'' in
  \emph{NeurIPS}, 2014.

\bibitem{yeh2016semantic}
R.~A. Yeh, Z.~Liu, D.~B. Goldman, and A.~Agarwala, ``Semantic facial expression
  editing using autoencoded flow,'' \emph{arXiv}, 2016.

\bibitem{ding2018exprgan_aaai}
H.~Ding, K.~Sricharan, and R.~Chellappa, ``Exprgan: Facial expression editing
  with controllable expression intensity,'' in \emph{AAAI}, 2018.

\bibitem{park2020swapping_nips}
T.~Park, J.-Y. Zhu, O.~Wang, J.~Lu, E.~Shechtman, A.~A. Efros, and R.~Zhang,
  ``Swapping autoencoder for deep image manipulation,'' in \emph{NeurIPS},
  2020.

\bibitem{hu2020unsupervised}
B.~Hu, Z.~Zheng, P.~Liu, W.~Yang, and M.~Ren, ``Unsupervised eyeglasses removal
  in the wild,'' \emph{IEEE Transactions on Cybernetics}, 2020.

\bibitem{shen2020interpreting_cvpr}
Y.~Shen, J.~Gu, X.~Tang, and B.~Zhou, ``Interpreting the latent space of gans
  for semantic face editing,'' in \emph{CVPR}, 2020.

\bibitem{Cozzolino2017}
D.~Cozzolino, G.~Poggi, and L.~Verdoliva, ``{Recasting residual-based local
  descriptors as convolutional neural networks: An application to image forgery
  detection},'' in \emph{IH and MMSec 2017 - Proceedings of the ACM Workshop on
  Information Hiding and Multimedia Security}, 2017.

\bibitem{fridrich2012rich}
J.~Fridrich and J.~Kodovsky, ``Rich models for steganalysis of digital
  images,'' \emph{IEEE Transactions on Information Forensics and Security},
  2012.

\bibitem{Afchar2018}
D.~Afchar, V.~Nozick, J.~Yamagishi, and I.~Echizen, ``{MesoNet: A compact
  facial video forgery detection network},'' in \emph{2018 IEEE International
  Workshop on Information Forensics and Security (WIFS)}, 2018.

\bibitem{Zhou2017}
P.~Zhou, X.~Han, V.~I. Morariu, and L.~S. Davis, ``{Two-Stream Neural Networks
  for Tampered Face Detection},'' in \emph{ICCVW}, 2017.

\bibitem{Nguyen2019}
H.~H. Nguyen, J.~Yamagishi, and I.~Echizen, ``{Capsule-forensics: Using Capsule
  Networks to Detect Forged Images and Videos},'' in \emph{ICASSP}, 2019.

\bibitem{szegedy2015going_cvpr}
C.~Szegedy, W.~Liu, Y.~Jia, P.~Sermanet, S.~Reed, D.~Anguelov, D.~Erhan,
  V.~Vanhoucke, and A.~Rabinovich, ``Going deeper with convolutions,'' in
  \emph{CVPR}, 2015.

\bibitem{szegedy2016rethinking}
C.~Szegedy, V.~Vanhoucke, S.~Ioffe, J.~Shlens, and Z.~Wojna, ``Rethinking the
  inception architecture for computer vision,'' in \emph{CVPR}, 2016.

\bibitem{Sabour2017}
S.~Sabour, N.~Frosst, and G.~E. Hinton, ``{Dynamic routing between capsules},''
  in \emph{NeurIPS}, 2017.

\bibitem{Wang2020}
K.~Wang, X.~Peng, J.~Yang, D.~Meng, and Y.~Qiao, ``{Region Attention Networks
  for Pose and Occlusion Robust Facial Expression Recognition},'' \emph{IEEE
  Transactions on Image Processing}, 2020.

\bibitem{Li2020}
S.~Li and W.~Deng, ``{A Deeper Look at Facial Expression Dataset Bias},''
  \emph{IEEE Transactions on Affective Computing}, 2020.

\bibitem{Nirkin2021}
Y.~Nirkin, L.~Wolf, Y.~Keller, and T.~Hassner, ``{DeepFake Detection Based on
  the Discrepancy Between the Face and its Context},'' \emph{IEEE Transactions
  on Pattern Analysis and Machine Intelligence}, 2021.

\bibitem{Qian2020_eccv2020}
Y.~Qian, G.~Yin, L.~Sheng, Z.~Chen, and J.~Shao, ``{Thinking in Frequency: Face
  Forgery Detection by Mining Frequency-aware Clues},'' in \emph{ECCV}, 2020.

\bibitem{Shen2021_aaai2021}
C.~Shen, Y.~Taiping, C.~Yang, D.~Shouhong, L.~Jilin, and R.~Ji, ``{Local
  Relation Learning for Face Forgery Detection},'' in \emph{AAAI}, 2021.

\bibitem{Li2021_cvpr2021}
J.~Li, H.~Xie, J.~Li, Z.~Wang, and Y.~Zhang, ``{Frequency-aware Discriminative
  Feature Learning Supervised by Single-Center Loss for Face Forgery
  Detection},'' in \emph{CVPR}, 2021.

\bibitem{Frank2020_icml2020}
J.~Frank, T.~Eisenhofer, L.~Sch{\"{o}}nherr, A.~Fischer, D.~Kolossa, and
  T.~Holz, ``{Leveraging frequency analysis for deep fake image recognition},''
  in \emph{ICML}, 2020.

\bibitem{Liu2021_cvpr2021}
H.~Liu, X.~Li, W.~Zhou, Y.~Chen, Y.~He, H.~Xue, W.~Zhang, and N.~Yu,
  ``{Spatial-Phase Shallow Learning: Rethinking Face Forgery Detection in
  Frequency Domain},'' in \emph{CVPR}, 2021.

\bibitem{Mirsky2020}
Y.~Mirsky and W.~Lee, ``{The Creation and Detection of Deepfakes: A Survey},''
  \emph{ACM Computing Surveys (CSUR)}, 2020.

\bibitem{FaceSwap}
``https://github.com/marekkowalski/faceswap,'' 2018.

\bibitem{Thies2019}
J.~Thies, M.~Zollh{\"{o}}fer, and M.~Nie{\ss}ner, ``{Deferred neural rendering:
  Image synthesis using neural textures},'' in \emph{siggraph}, 2019.

\bibitem{he2018amc_eccv}
Y.~He, J.~Lin, Z.~Liu, H.~Wang, L.-J. Li, and S.~Han, ``Amc: Automl for model
  compression and acceleration on mobile devices,'' in \emph{ECCV}, 2018.

\bibitem{cubuk2019autoaugment_cvpr}
E.~D. Cubuk, B.~Zoph, D.~Mane, V.~Vasudevan, and Q.~V. Le, ``Autoaugment:
  Learning augmentation strategies from data,'' in \emph{CVPR}, 2019.

\bibitem{real2019regularized}
E.~Real, A.~Aggarwal, Y.~Huang, and Q.~V. Le, ``Regularized evolution for image
  classifier architecture search,'' in \emph{AAAI}, 2019.

\bibitem{liu2019auto}
C.~Liu, L.-C. Chen, F.~Schroff, H.~Adam, W.~Hua, A.~L. Yuille, and L.~Fei-Fei,
  ``Auto-deeplab: Hierarchical neural architecture search for semantic image
  segmentation,'' in \emph{CVPR}, 2019.

\bibitem{ghiasi2019fpn}
G.~Ghiasi, T.-Y. Lin, and Q.~V. Le, ``Nas-fpn: Learning scalable feature
  pyramid architecture for object detection,'' in \emph{CVPR}, 2019.

\bibitem{zhang2020efficientpose}
W.~Zhang, J.~Fang, X.~Wang, and W.~Liu, ``Efficientpose: Efficient human pose
  estimation with neural architecture search,'' \emph{arXiv}, 2020.

\bibitem{song2020efficient}
D.~Song, C.~Xu, X.~Jia, Y.~Chen, C.~Xu, and Y.~Wang, ``Efficient residual dense
  block search for image super-resolution,'' in \emph{AAAI}, 2020.

\bibitem{dong2019one_iccv2019}
X.~Dong and Y.~Yang, ``One-shot neural architecture search via self-evaluated
  template network,'' in \emph{ICCV}, 2019.

\bibitem{dong2019network_neurips2019}
------, ``Network pruning via transformable architecture search,'' in
  \emph{NeurIPS}, 2019.

\bibitem{He_2020_CVPR}
Y.~He, Y.~Ding, P.~Liu, L.~Zhu, H.~Zhang, and Y.~Yang, ``Learning filter
  pruning criteria for deep convolutional neural networks acceleration,'' in
  \emph{CVPR}, 2020.

\bibitem{zoph2018learning}
B.~Zoph, V.~Vasudevan, J.~Shlens, and Q.~V. Le, ``Learning transferable
  architectures for scalable image recognition,'' in \emph{CVPR}, 2018.

\bibitem{afchar2018mesonet}
D.~Afchar, V.~Nozick, J.~Yamagishi, and I.~Echizen, ``Mesonet: a compact facial
  video forgery detection network,'' in \emph{2018 IEEE International Workshop
  on Information Forensics and Security (WIFS)}, 2018.

\bibitem{Qian2020}
Y.~Qian, G.~Yin, L.~Sheng, Z.~Chen, and J.~Shao, ``{Thinking in Frequency: Face
  Forgery Detection by Mining Frequency-aware Clues},'' in \emph{ECCV}, 2020.

\bibitem{Liu2014}
P.~Liu, S.~Han, Z.~Meng, and Y.~Tong, ``{Facial expression recognition via a
  boosted deep belief network},'' in \emph{CVPR}, 2014.

\bibitem{Zhang2020_ijcai}
H.~Zhang, W.~Su, and Z.~Wang, ``{Weakly Supervised Local-Global Attention
  Network for Facial Expression Recognition},'' in \emph{IJCAI}, 2020.

\bibitem{li2020face_cvpr2020}
L.~Li, J.~Bao, T.~Zhang, H.~Yang, D.~Chen, F.~Wen, and B.~Guo, ``Face x-ray for
  more general face forgery detection,'' in \emph{CVPR}, 2020.

\bibitem{Deepfakes}
``https://github.com/deepfakes/faceswap,'' 2018.

\bibitem{MatthiasNiessner2016}
{Matthias Niessner}, ``{Face2Face: Real-time Face Capture and Reenactment of
  RGB Videos (CVPR 2016 Oral) - YouTube},'' in \emph{CVPR}, 2016.

\bibitem{bayar2016deep}
B.~Bayar and M.~C. Stamm, ``A deep learning approach to universal image
  manipulation detection using a new convolutional layer,'' in
  \emph{Proceedings of the 4th ACM Workshop on Information Hiding and
  Multimedia Security}, 2016.

\bibitem{cozzolino2017recasting}
D.~Cozzolino, G.~Poggi, and L.~Verdoliva, ``Recasting residual-based local
  descriptors as convolutional neural networks: an application to image forgery
  detection,'' in \emph{Proceedings of the 5th ACM Workshop on Information
  Hiding and Multimedia Security}, 2017.

\bibitem{rahmouni2017distinguishing}
N.~Rahmouni, V.~Nozick, J.~Yamagishi, and I.~Echizen, ``Distinguishing computer
  graphics from natural images using convolution neural networks,'' in
  \emph{2017 IEEE Workshop on Information Forensics and Security (WIFS)}, 2017.

\bibitem{Nguyen2019_bats2019}
H.~H. Nguyen, F.~Fang, J.~Yamagishi, and I.~Echizen, ``{Multi-task Learning for
  Detecting and Segmenting Manipulated Facial Images and Videos},'' in
  \emph{BTAS}, 2019.

\bibitem{Hussain2021_wacv}
S.~Hussain, P.~Neekhara, M.~Jere, F.~Koushanfar, and J.~McAuley, ``{Adversarial
  Deepfakes: Evaluating Vulnerability of Deepfake Detectors to Adversarial
  Examples},'' in \emph{WACV}, 2021.

\bibitem{simonyan2014very}
K.~Simonyan and A.~Zisserman, ``Very deep convolutional networks for
  large-scale image recognition,'' in \emph{ICLR}, 2015.

\bibitem{8638330_wacv2019}
F.~Matern, C.~Riess, and M.~Stamminger, ``Exploiting visual artifacts to expose
  deepfakes and face manipulations,'' in \emph{WACV}, 2019.

\bibitem{Li2020_mm2020}
X.~Li, Y.~Lang, Y.~Chen, X.~Mao, Y.~He, S.~Wang, H.~Xue, and Q.~Lu, ``{Sharp
  Multiple Instance Learning for DeepFake Video Detection},'' in \emph{ACM MM},
  2020.

\bibitem{Houlden2017_nips2017}
R.~L. Houlden, S.~Moore, W.~Cornish, and K.~Tiwana, ``{Train longer, generalize
  better: closing the generalization gap in large batch training of neural
  networks},'' in \emph{NeurIPS}, 2017.

\bibitem{Peng_2017_CVPR}
C.~Peng, X.~Zhang, G.~Yu, G.~Luo, and J.~Sun, ``Large kernel matters -- improve
  semantic segmentation by global convolutional network,'' in \emph{CVPR},
  2017.

\bibitem{Li_2020_CVPR}
L.~Li, J.~Bao, H.~Yang, D.~Chen, and F.~Wen, ``Advancing high fidelity identity
  swapping for forgery detection,'' in \emph{CVPR}, 2020.

\bibitem{Wang2021}
Y.~Wang, X.~Chen, J.~Zhu, W.~Chu, Y.~Tai, C.~Wang, J.~Li, Y.~Wu, F.~Huang, and
  R.~Ji, ``{HifiFace: 3D Shape and Semantic Prior Guided High Fidelity Face
  Swapping},'' in \emph{CVPR}, 2021.

\bibitem{zhang2021prouigan_arxiv2021}
Y.~Zhang, X.~Yu, X.~Lu, and P.~Liu, ``Pro-uigan: Progressive face hallucination
  from occluded thumbnails,'' \emph{arXiv}, 2021.

\bibitem{qi2020deeprhythm_acmmm_2020}
H.~Qi, Q.~Guo, F.~Juefei-Xu, X.~Xie, L.~Ma, W.~Feng, Y.~Liu, and J.~Zhao,
  ``Deeprhythm: Exposing deepfakes with attentional visual heartbeat rhythms,''
  in \emph{ACM MM}, 2020.

\bibitem{wang2020deepsonar_acmmm2020}
R.~Wang, F.~Juefei-Xu, Y.~Huang, Q.~Guo, X.~Xie, L.~Ma, and Y.~Liu,
  ``Deepsonar: Towards effective and robust detection of ai-synthesized fake
  voices,'' in \emph{ACM MM}, 2020.

\end{thebibliography}

\end{document}